\documentclass[acmtog]{acmart}

\usepackage{booktabs}
\usepackage{graphicx}
\usepackage{amsmath}
\usepackage{booktabs}
\usepackage{multirow}
\usepackage{colortbl}
\usepackage{makecell}
\usepackage{diagbox}
\usepackage[capitalize]{cleveref}

\crefname{section}{Sec.}{Secs.}
\Crefname{section}{Section}{Sections}
\crefname{table}{Tab.}{Tabs.}
\Crefname{table}{Table}{Tables}
\crefname{figure}{Fig.}{Figs.}
\Crefname{figure}{Figure}{Figures}
\crefname{equation}{Eq.}{Eqs.}
\Crefname{equation}{Equation}{Equations}

\newcommand{\supp}{\textit{Supplemental Material}\xspace}

\newcommand{\R}[1]{{%
      \textbf{%
        \ifstrequal{#1}{1}{\textcolor{red}{R#1}}{%
          \ifstrequal{#1}{2}{\textcolor{blue}{R#1}}{%
            \ifstrequal{#1}{3}{\textcolor{magenta}{R#1}}{%
              \ifstrequal{#1}{4}{\textcolor{teal}{R#1}}{%
                \textcolor{cyan}{R#1}%
              }}}}%
      }%
    }}

\usepackage{xcolor}
\definecolor{myred}{rgb}{0.8,0,0}
\definecolor{mygreen}{rgb}{0,0.8,0}
\definecolor{myblue}{rgb}{0,0,0.95}
\definecolor{mypurple}{rgb}{0.75,0,0.75}
\definecolor{myorange}{rgb}{0.75,0.25,0.25}
\usepackage{pifont}

\newcommand{\camera}[1]{{#1}} %
\newcommand{\revision}[1]{{#1}} %

\newcommand{\method}{\texttt{TGH}\xspace}
\newcommand{\tgh}{Temporal Gaussian Hierarchy\xspace}
\newcommand{\cpm}{Compact Appearance Model\xspace}

\usepackage[ruled]{algorithm2e}

\SetAlFnt{\small}
\SetAlCapFnt{\small}
\SetAlCapNameFnt{\small}
\SetAlCapHSkip{0pt}

\AtBeginDocument{%
\providecommand\BibTeX{{%
\normalfont B\kern-0.5em{\scshape i\kern-0.25em b}\kern-0.8em\TeX}}}

\acmSubmissionID{415}

\setcopyright{acmlicensed}
\acmJournal{TOG}
\acmYear{2024} \acmVolume{43} \acmNumber{6} \acmArticle{171}
\acmMonth{12}\acmDOI{10.1145/3687919}

\acmSubmissionID{415}

\title{Representing Long Volumetric Video with Temporal Gaussian Hierarchy}

\author{Zhen Xu}
\orcid{0009-0002-6098-4198}
\email{zhenx@zju.edu.cn}
\affiliation{
    \institution{State Key Lab of CAD\&CG, Zhejiang University}
    \city{Hangzhou}
    \country{China}
}
\authornote{Denotes equal contribution.}

\author{Yinghao Xu}
\orcid{0000-0003-2696-9664}
\email{justimyhxu@gmail.com}
\affiliation{
    \institution{Stanford University}
    \city{Stanford}
    \country{United States}
}
\authornotemark[1]

\author{Zhiyuan Yu}
\orcid{0000-0001-7220-7789}
\email{zyuaq@ust.hk}
\affiliation{
    \institution{Hong Kong University of Science and Technology}
    \city{Hong Kong}
    \country{China}
}
\authornotemark[1]

\author{Sida Peng}
\orcid{0000-0001-6546-4525}
\email{pengsida@zju.edu.cn}
\affiliation{
    \institution{Zhejiang University}
    \city{Hangzhou}
    \country{China}
}

\author{Jiaming Sun}
\orcid{0000-0003-4053-8510}
\email{suenjiaming@gmail.com}
\affiliation{
    \institution{Zhejiang University}
    \city{Hangzhou}
    \country{China}
}

\author{Hujun Bao}
\orcid{0000-0002-2662-0334}
\email{bao@cad.zju.edu.cn}
\affiliation{
    \institution{State Key Lab of CAD\&CG, Zhejiang University}
    \city{Hangzhou}
    \country{China}
}

\author{Xiaowei Zhou}
\orcid{0000-0003-1926-5597}
\email{xwzhou@zju.edu.cn}
\affiliation{
    \institution{State Key Lab of CAD\&CG, Zhejiang University}
    \city{Hangzhou}
    \country{China}
}
\authornote{
    Corresponding author: Xiaowei Zhou.
}

\begin{document}

\begin{abstract}

    This paper aims to address the challenge of reconstructing long volumetric videos from multi-view RGB videos.
    Recent dynamic view synthesis methods leverage powerful 4D representations, like feature grids or point cloud sequences, to achieve high-quality rendering results. However, they are typically limited to short (1$\sim$2s) video clips and often suffer from large memory footprints when dealing with longer videos.
    To solve this issue, we propose a novel 4D representation, named Temporal Gaussian Hierarchy, to compactly model long volumetric videos.
    Our key observation is that there are generally various degrees of temporal redundancy in dynamic scenes, which consist of areas changing at different speeds.
    Motivated by this, our approach builds a multi-level hierarchy of 4D Gaussian primitives, where each level separately describes scene regions with different degrees of content change, and adaptively shares Gaussian primitives to represent unchanged scene content over different temporal segments, thus effectively reducing the number of Gaussian primitives.
    In addition, the tree-like structure of the Gaussian hierarchy allows us to efficiently represent the scene at a particular moment with a subset of Gaussian primitives, leading to nearly constant GPU memory usage during the training or rendering regardless of the video length.
    Moreover, we design a \cpm that mixes diffuse and view-dependent Gaussians to further minimize the model size while maintaining the rendering quality.
    We also develop a rasterization pipeline of Gaussian primitives based on the hardware-accelerated technique to improve rendering speed.
    Extensive experimental results demonstrate the superiority of our method over alternative methods in terms of training cost, rendering speed, and storage usage.
    To our knowledge, this work is the first approach capable of efficiently handling minutes of volumetric video data while maintaining state-of-the-art rendering quality.
\end{abstract}



\begin{teaserfigure}
    \centering
    \includegraphics[width=1\textwidth]{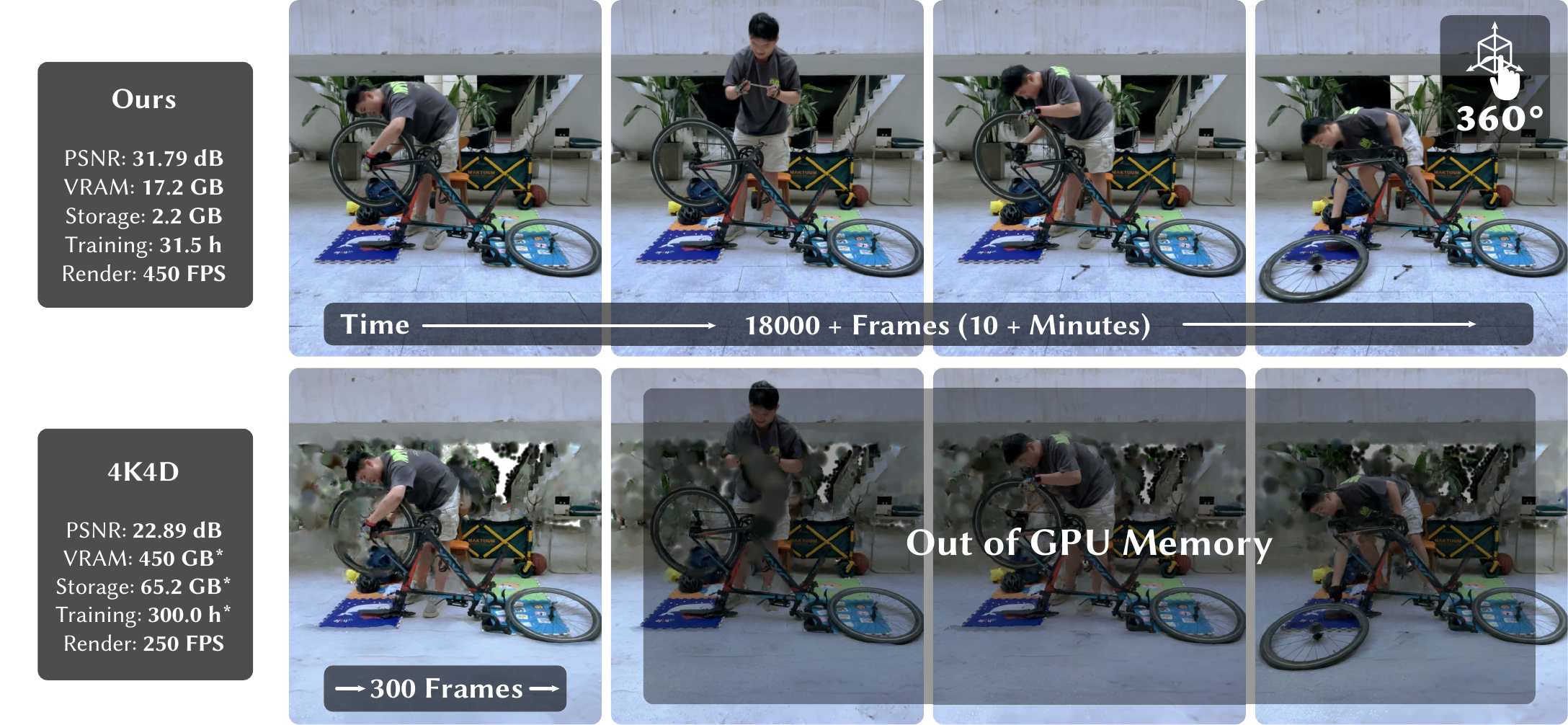}
    \caption{\textbf{Photorealistic rendering of a long volumetric video with 18,000 frames}. Our proposed method utilizes an efficient 4D representation with Temporal Gaussian Hierarchy, requiring only 17.2 GB of VRAM and 2.2 GB of storage for 18,000 frames. This achieves a 30x and 26x reduction compared to the previous state-of-the-art 4K4D method~\cite{xu20244k4d}.
        Notably, 4K4D~\cite{xu20244k4d} could only handle 300 frames with a 24GB RTX 4090 GPU, whereas our method can process the entire 18,000 frames, thanks to the constant computational cost enabled by our \tgh.
        Our method supports real-time rendering at 1080p resolution with a speed of 450 FPS using an RTX 4090 GPU while maintaining state-of-the-art quality.}
    \label{fig:teaser}
\end{teaserfigure}

\maketitle

\begin{figure}
    \includegraphics[width=1\linewidth]{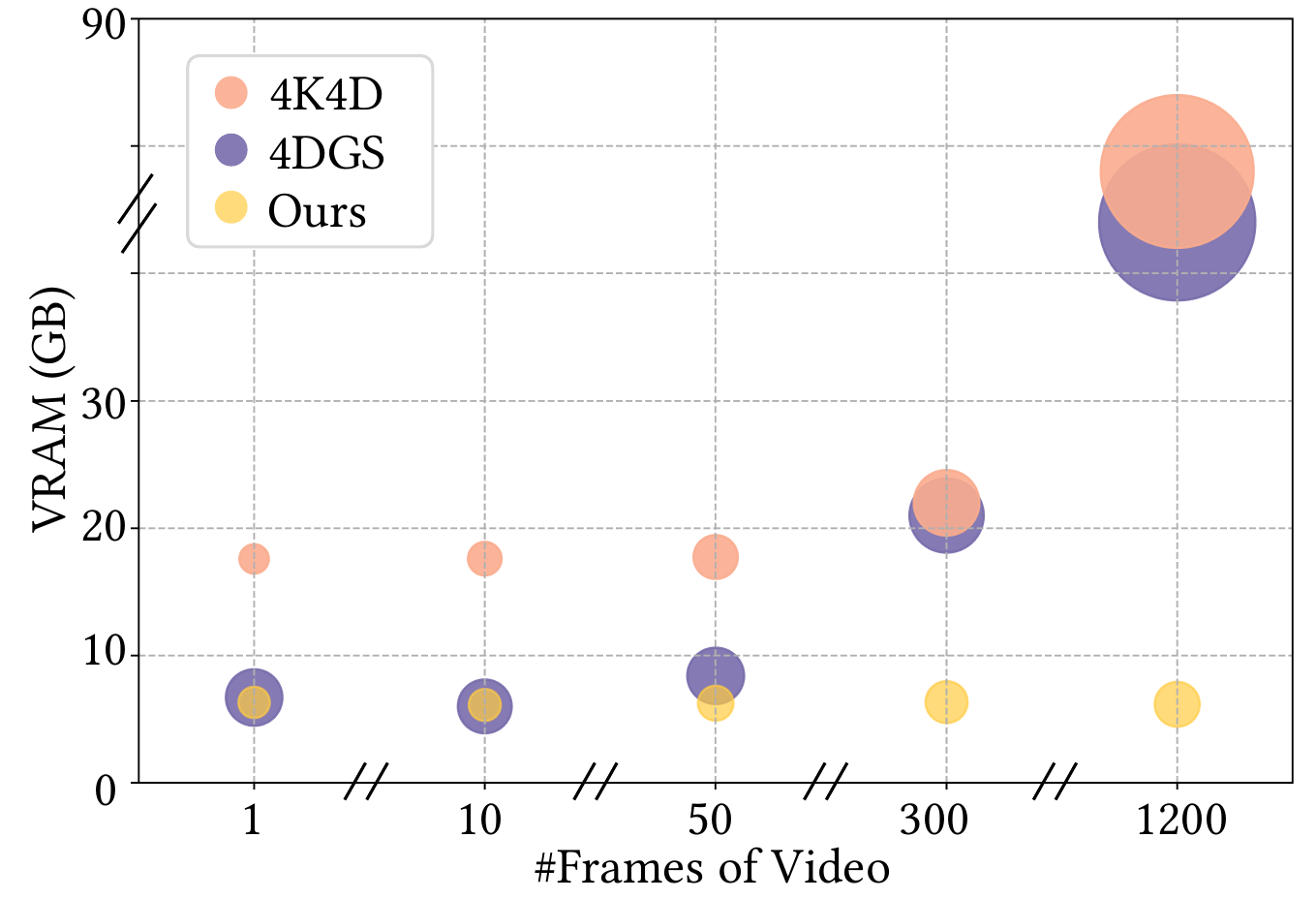}
    \caption{\textbf{Training cost v.s. number of video frames}. By varying the number of video frames, we compare our method with recent state-of-the-art volumetric video techniques on the Neural3DV~\cite{li2022neural} dataset in terms of training cost and storage usage, measured using VRAM (GB) and model size (MB), respectively.
        Each bubble’s area is proportional to its storage usage.
        Our method consistently maintains a constant training cost and near-constant storage usage regardless of the video frame length, demonstrating the scalability of our method for very long volumetric videos.}
    \label{fig:teaser-plot}
\end{figure}

\section{Introduction}

Volumetric videos aim to capture dynamic scenes from multiple viewpoints and provide the capability for free-viewpoint synthesis, enabling users to interact with virtual objects or characters for an immersive experience.
The availability of high-quality volumetric videos is crucial in many domains including AR/VR, gaming, telepresence, and others.
Although traditional volumetric video systems \cite{collet2015hqfvv,wu2020nhr} have shown impressive results, they typically require bespoke hardware and sophisticated studio setup, which limits their accessibility and practicality.

Emerging neural rendering methods offer the ability to perform photorealistic view synthesis of dynamic scenes based on implicit or explicit 4D representations.
Early work \cite{pumarola2021dnerf,li2022neural} aims to utilize neural radiance fields with temporal embeddings to parameterize dynamic scenes.
Despite its storage-friendly model size, this representation has limited representation power and efficiency, leading to low-quality renderings that take several seconds or even minutes per frame.
Recent research \cite{kplanes_2023,Cao2023HEXPLANE,Wang_2022_CVPR,xu20244k4d} employs more powerful 4D representations, either through feature grids or point cloud sequences, to enhance rendering fidelity or efficiency.
However, these methods generally only operate on short volumetric video sequences (1$\sim$2s)~\cite{xu20244k4d, yang2023realtime}.
When applied to longer videos (1$\sim$10minutes), these approaches require very large models, posing scaling challenges considering practical storage size constraints.

In this paper, we introduce Temporal Gaussian Hierarchy, a novel 4D representation for compact modeling of long volumetric videos, while being efficient in training and rendering.
Our key observation is that dynamic scenes generally consist of areas that change slowly and areas that change quickly, reflecting varying degrees of temporal redundancy.
Motivated by this observation, our approach builds a multi-level hierarchy of scene primitives, e.g., 3D or 4D Gaussian splats \cite{kerbl20233d,yang2023realtime}, to represent 4D scenes, where different levels model areas with different degrees of content change, and assign different numbers of scene primitives to each level, aiming to handle the temporal redundancy explicitly.
Specifically, as illustrated in \cref{fig:framework}, the proposed hierarchical representation consists of multiple levels, and each level contains a set of segments responsible for modeling scenes within different temporal scales.
Within each segment, we utilize a set of 4D Gaussian primitives \cite{yang2023realtime} to represent the scene, considering their high representation capability and rendering speed.
Our experimental results demonstrate that in the proposed hierarchy, scene areas with slower motions generally share Gaussian primitives within segments across longer temporal scales, thereby reducing the number of scene primitives and leading to a decrease in model size.
Moreover, the temporal Gaussian hierarchy allows us to represent the scene at a time step with a subset of segments instead of all Gaussian primitives, thus further reducing runtime memory.

Another advantage of temporal Gaussian hierarchy is that it enables us to keep efficient training and rendering regardless of the length of volumetric videos.
The tree-like structure of our representation can efficiently locate the segments corresponding to a particular moment.
Therefore, we can store the Gaussian hierarchy in random-access memory, and load the corresponding blocks into the GPU memory during the training or rendering of a scene at a specific moment.
This strategy ensures that the GPU memory usage during runtime remains constant, regardless of the length of the volumetric video, guaranteeing efficient training and rendering.

To further reduce the model size of the volumetric video, we introduce a hybrid appearance model that consists of diffuse Gaussians and view-dependent Gaussians, which adaptively assigns the representational power to different areas of target scenes.
This strategy can also mitigate the overfitting problems of Gaussian primitives while preserving the high capability of modeling dynamic view-dependent appearance.
Lastly, we develop a GPU-accelerated algorithm and hardware-accelerated rasterization pipeline for our representation to greatly enhance the \textbf{rendering speed}.
As illustrated in \cref{fig:teaser}, our method significantly surpasses previous methods in terms of training cost, rendering efficiency, and storage overhead, while achieving superior rendering quality.

We evaluate our method across various widely used datasets for dynamic view synthesis given multi-view input.
Additionally, we collect a set of long multi-view sequences (5-10 minutes) to demonstrate the scalability of our method on video length.
Both qualitative and quantitative results demonstrate that our method outperforms existing baselines, showcasing state-of-the-art visual quality and rendering speed while requiring significantly less training cost and memory usage, enabling the reconstruction and rendering of long volumetric videos.
Using an RTX 4090 GPU, our method can process videos with 18,000 frames at a 1080p resolution, achieving a rendering speed of 450 FPS. We also collect a long real-world multi-view dataset, dubbed \textit{SelfCap}, to validate our method.

Our contributions can be summarized as follows:
\begin{itemize}
    \item We introduce a novel, efficient, and expressive \tgh representation for long volumetric video. To our knowledge, our method is the first approach capable of handling minutes of volumetric video data.
    \item We propose a \cpm and a new rasterization implementation to facilitate real-time and high-quality dynamic view synthesis while maintaining a compact size.
    \item \revision{We propose a system to efficiently model long volumetric videos for the first time and demonstrate} state-of-the-art dynamic view synthesis quality on the \textit{Neural3DV} \cite{li2022neural}, \textit{ENeRF-Outdoor \cite{lin2022efficient}} and \textit{MobileStage} \cite{xu20244k4d} datasets, while also achieving the best rendering speed with reduced training cost and memory usage.
\end{itemize}

\section{Related Work}

\subsection{Volumetric Videos}
Volumetric videos capture dynamic scenes from multiple viewpoints and provide the capability for immersive free-viewpoint synthesis. Earlier works leverage template-priors like  shape-from-silhouettes~\cite{ahmed2008dense}, deformable models~\cite{carranza2003free} and template-free dynamic reconstruction systems with depth sensor integration~\cite{newcombe2015dynamicfusion}.
Some research explores fusion methodologies, incorporating priors like skeletal structures and parametric body shapes to facilitate non-rigid 3D reconstruction, as seen in the works of ~\cite{yu2018doublefusion, yu2017bodyfusion}.
Another research line explores data-driven approaches~\cite{bozic2020deepdeform} for dynamic 3D reconstruction beyond photometric consistency criteria, with works like ~\cite{lombardi2019neural} introducing generative models for more flexible reconstruction frameworks.
These methods either cannot be applied to complex scenes with many occlusions, require parameterized templates, or rely on expensive and hardware-intensive capture devices, all of which limit the application of volumetric video in real-world scenarios.

\subsection{Neural Scene Representations for Dynamic Scenes}

Neural scene representations have attracted significant attention from the research community due to their impressive performance in 3D reconstruction and novel-view synthesis.
Methods such as NeRF~\cite{mildenhall2021nerf}, SDF~\cite{park2019deepsdf}, and OCC-Net~\cite{occnet} parameterize scenes as radiance fields, signed distance functions, and occupancy using neural networks.
However, these techniques struggle to generalize to dynamic scenes, because of the illumination variations and complex object motions.

Recent advancements~\cite{fang2022fast, park2021hypernerf, park2021nerfies, pumarola2021d, tretschk2021non} have seen efforts to extend NeRF into volumetric videos for dynamic view synthesis.
Some approaches model dynamic scenes~\cite{park2021hypernerf, park2021nerfies, pumarola2021d} by learning a canonical static template with deformable fields represented by spatial offsets.
Yet, these methods can only handle limited camera changes and object motions.
Some approaches~\cite{du2021neural} directly employ 4D NeRF, incorporating timestamps along with spatial location and view direction to address spatial changes.
To enhance rendering quality, other works explore hybrid representations~\cite{kplanes_2023, Cao2023HEXPLANE, xu20244k4d} to parameterize dynamic scenes.
However, these approaches often incur substantial computational costs for training and storage, limiting their practicality and quality.
Alternatively, some methods~\cite{yu2021pixelnerf, chen2021mvsnerf, long2022sparseneus, wang2021ibrnet, li2023dynibar} aim to construct feed-forward models for inferring 3D NeRF from multi-view images, thereby extending their applicability to dynamic scenes.
However, due to the lack of temporal modeling and the computational overhead of convolution operations in image encoders , these methods suffers from the low-quality renderings and slow rendering speeds, especially for long volumetric videos.

To accelerate NeRF rendering, several approaches utilize explicit structures, employing shallow MLPs to represent scenes. These include voxel grids~\cite{garbin2021fastnerf, fridovich2022plenoxels, hedman2021baking, muller2022instant}, surfaces~\cite{lu2023urban, kulhanek2023tetra, chen2023mobilenerf, hasselgren2022shape}, and point-based representations~\cite{aliev2020neural, lassner2021pulsar, rakhimov2022npbg++, ruckert2022adop}. While these methods significantly reduce the computational load of neural network evaluations for rendering, they often incur additional storage consumption.
To mitigate memory costs, recent efforts have explored weight decomposition techniques~\cite{takikawa2022variable, tang2022compressible, chen2022tensorf}, while limiting their capacity to represent dynamic scenes effectively. Some methods~\cite{Wang_2022_CVPR,park2021hypernerf,tretschk2021nonrigid} have also proposed dynamic modeling on sparse voxel grids and deformable fields for videos, showing promise for short videos but introducing storage overhead when applied to longer volumetric sequences.
In contrast, our proposed approach introduces an efficient and compact \tgh for dynamic view synthesis.
This framework facilitates scalability to high-resolution and extensive volumetric videos while maintaining real-time rendering speeds.

\begin{figure*}
    \includegraphics[width=1.0\linewidth]{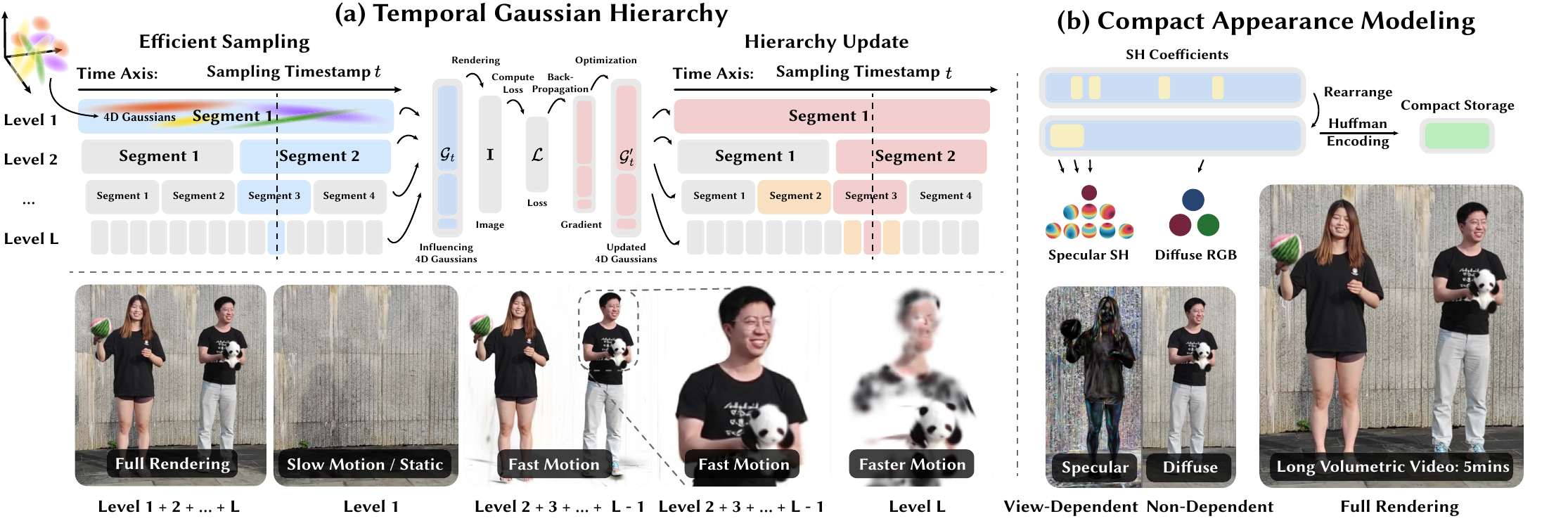}
    \caption{\textbf{Overview of our framework.} Given a long multi-view video sequence, our method can generate a compact volumetric video with minimal training and memory usage while maintaining real-time rendering with state-of-the-art quality. (a) We propose a hierarchical structure where each level consists of multiple temporal segments. Each segment stores a set of 4D Gaussians~\cite{yang2023realtime} to parametrize scenes. As shown at the bottom, the 4D Gaussians in different segments represent different granularities of motions, efficiently and effectively modeling video dynamics. (b) The appearance model leverages gradient thresholding to obtain sparse Spherical Harmonics coefficients, resulting in very compact storage while still maintaining view-dependent effects well.}
    \label{fig:framework}
\end{figure*}

\subsection{Gaussian Splatting for Dynamic Scenes}
3D Gaussians~\cite{kerbl20233d} have become widely popular due to their capacity for efficiently reconstructing high-fidelity 3D scenes from posed images.
In order to adapt 3D Gaussians for dynamic scenes, \camera{Dy3DGS~\cite{luiten2023dynamic} first proposes to reconstruct dynamic scenes by tracking 3D Gaussians on a frame-by-frame basis}. 
\camera{However, their approach leads to large models for long sequences since they need to store the per-point transform information for every frame.}
\camera{In comparison, our method takes advantage of the different degrees of temporal redundancy by using segments of varying lengths at different levels, sharing temporal information across frames to maintain a low disk storage overhead.}
DynamicGaussian~\cite{yang2023deformable, lu20243d} proposes learning the temporal motion and rotation of each 3D Gaussian, thereby supporting dynamic tracking applications.
Some studies utilize more powerful modules~\cite{wu20234d} like MLP to predict temporal motion or integrate dynamic Gaussians into monocular videos~\cite{yang2023deformable} or human avatar animations~\cite{jung2023deformable}. However, it is very challenging to parameterize very sudden motions on dynamic objects.

Another line of research leverages 4D Gaussians~\cite{yang2023realtime, duan20244d, lin2023gaussian} to model 3D dynamics.
These studies propose to learn high-dimensional rotations along the time axis and utilize marginal observations at specific timesteps to derive the 3D Gaussian for fast rendering.
Many works leverage this underlying representation for 4D generation~\cite{ren2023dreamgaussian4d, ling2023align}.
However, these frameworks incur significant computational overhead such as VRAM and storage when processing extensive video data.
For example, \camera{4DGS~\cite{yang2023realtime} requires to render a scene at a particular moment with all Gaussian primitives across the whole temporal scale, which can easily lead to the VRAM exhaustion}.
In contrast, our proposed \tgh allows using a particular set of 4D Gaussians~\cite{yang2023realtime} to describe the content of dynamic scenes within the corresponding segment.
Thanks to this design, our model maintains nearly constant memory costs for both training and inference, along with compact storage.
This feature enables our method to scale effectively to very long volumetric videos for dynamic view synthesis.

\section{Method}
\subsection{Overview}

Given multi-view videos, our goal is to efficiently reconstruct a corresponding compact and lightweight 4D volumetric representation that can be played back in real-time.
To achieve this goal, we first develop a novel 4D representation, dubbed \tgh, for efficiently modeling the different scales of motion and dynamics in a long dynamic scene (\cref{sec:temporal_gaussian_hierarchy}).
Then, we propose a compact appearance representation (\cref{sec:compact_appearance}), where only the few view-dependent points are modeled using all degrees of Spherical Harmonics \cite{muller2006spherical} to greatly reduce the storage cost for representing long volumetric videos while maintaining the rendering quality.
Finally, we develop an efficient rendering pipeline with hardware-accelerated rasterization and precomputation for rendering \method in real-time (\cref{sec:rendering}).
An overview of the proposed method can be found in \cref{fig:framework}.

\subsection{\tgh}
\label{sec:temporal_gaussian_hierarchy}

Previous methods for volumetric videos
\cite{yang2023realtime,xu20244k4d,lin2023im4d} require significant GPU memory and time, both of which typically scale with the duration of the video, limiting their applicability to representing long volumetric videos.
To overcome these problems, we design a novel 4D representation, \tgh.
By leveraging a tree-like structure to manage hierarchical temporal segments for different motion granularities, we can efficiently identify all relevant segments for a specific timestamp and thus maintain a nearly-constant computation cost (VRAM and speed) for volumetric videos of arbitrary length.

\subsubsection{Temporal Hierarchy Structure}
\cref{fig:framework}(a) provides an structural overview of the \tgh.
Dynamic scenes typically encompass regions with varying scales of motion.
We represent this diversity using \tgh.
Concretely, we construct a hierarchical structure in which each level accommodates many temporal segments of equal length.
The segment length at deeper levels is half that of the preceding layer.
We denote the total duration of the video as $T$, the depth of hierarchy structure is $L$ and the segment length at the root level as $S$.
So the temporal scale of segments $s_l$ and the number of segments $N_{l}$ at level $l$ can be expressed as:
\begin{align}
     & s_l = \frac{S}{2^l},                    \\
     & N_{l} = \mathrm{ceil}(\frac{T}{s_{l}}),
\end{align}
where $\mathrm{ceil}$ denotes rounding up to the closest integer.
\camera{In addtion, we append a single global segment \revision{with a length of $\infty$ to capture static part of the scene.}}

In each segment, we store a set of 4D Gaussians (4DGS) \cite{yang2023realtime} to represent the dynamic scenes.
Specifically, 4D Gaussian is defined with 4D means $\boldsymbol{\mu} \in \mathbb{R}^4$, 4D scaling $\mathbf{s} \in \mathbb{R}^4$, a scalar opacity $o \in \mathbb{R}$ and two isotropic left and right quaternions $\mathbf{q}_l \in \mathbb{R}^4, \mathbf{q}_r \in \mathbb{R}^4$, base color $\mathbf{c}_{base} \in \mathbb{R}^3$, and residual spherical harmonics (SH) coefficients $\mathbf{h} \in \mathbb{R}^{m}$ with $m$ denoting the number of SH bases.
The 4D covariance matrix of the Gaussian distribution is defined as:
\begin{equation}
    \mathbf{\Sigma} = \mathbf{R} \mathbf{S} \mathbf{S}^T \mathbf{R}^T,
\end{equation}
where $\mathbf{S}$ is a diagonal matrix with $\mathbf{s}$ as its diagonal elements, and $\mathbf{R}$ is the symmetric expansion of the two isotropic quaternions.
The opacity of the 4D Gaussians follows a Gaussian distribution $o_t \sim \mathcal{N}(t; \mu_t, \sigma_t)$ around its temporal center $\mu_t$, with a scale characterized by the last element of the 4D covariance matrix $\sigma_t = \Sigma_{4,4}$ \camera{\cite{yang2023realtime}}.
Considering the temporal range where the Gaussian opacity is larger than a small value $o_{th}$,
we can compute the influence radius $r$ and influence range $[\overline{\tau},\underline{\tau}]$ of the Gaussian by inversely evaluating the function of the Gaussian distribution $\mathcal{N}(t;\mu_t,\sigma_t)$ using:

\begin{equation}
    \label{eq:gaussian_influence_range}
    \begin{aligned}
         & r = \mathrm{sqrt}(\frac{\mathrm{log}(o_{th})}{-0.5} \cdot \sigma_t),         \\
         & \overline{\tau} = \camera{\mu_t} - r, \underline{\tau} = \camera{\mu_t} + r.
    \end{aligned}
\end{equation}

\tgh is designed to model video dynamics efficiently and effectively: long segments at a coarse level work well for slow motions, while short segments at a finer level are better for fast motions.
To achieve this, when placing the 4D Gaussians in the hierarchical structure, we need to ensure the range of the segment covers the Gaussians influence radius within it and can't be covered by any segments in next levels.
Specifically, we first calculate the start $\overline{\tau}_{l,n}$ and end $\underline{\tau}_{l,n}$ timestamp of the $n$-th segment at level $l$ with temporal scale $s_l$:
\begin{equation}
    \label{eq:node_influence_range}
    \begin{aligned}
         & \overline{\tau}_{l,n} = n \cdot s_{l}, \textit{\ \ } \underline{\tau}_{l,n} = (n+1) \cdot s_{l},
    \end{aligned}
\end{equation}
and then use the following strategy to place the 4D Gaussians in the structure:
\begin{equation}
    \label{eq:gaussian_node_assignment}
    \begin{aligned}
                                              & \overline{\tau}_{l+0,n}    \leq  \overline{\tau}  \leq           \underline{\tau}  \leq \underline{\tau}_{l+0,n}, \\
        \forall m \in \{1,2...M\}, \textit{ } & \overline{\tau}_{l+1,m}  >     \overline{\tau}  \lor  \underline{\tau}  >    \underline{\tau}_{l+1,m},
    \end{aligned}
\end{equation}
where $m \in \{1,2...M\}$ denotes all segment indices and $M$ denotes the number of segments in the next level $l+1$.
This makes the \tgh structure unique and make sure that each 4D Gaussian is placed in the shortest possible segment.

\subsubsection{Efficient Rendering}
With the \tgh, we can efficiently sample the 4D Gaussians that influence this timestamp $t$ to render images.
Since the segments at each level don't overlap, the timestamp is influenced by only one segment at level $l$, and the corresponding index $n^{t}_{l}$ is:
\begin{align}
    \label{eq:efficient_sampling}
    n^{t}_{l} = \mathrm{floor}(\frac{t}{s_{l}}),
\end{align}
resulting in $L$ segments across all levels.
The computational complexity of this identification process is $O(\log N)$, where $N$ is the total number of segments in the hierarchy.
We then concatenate all 4D Gaussians within the influenced segments with index $\{ {n^{t}_{l}} | l \in {1...L}\}$ to obtain the streaming Gaussians $\mathcal{G}_{t}$ and render the images with a differentiable rasterizer.

In practice, we place the \tgh structure inside RAM and only copy the influenced segments to the GPU memory.
This helps save a significant amount of GPU memory and maintains near-constant GPU memory usage regardless of the length of the volumetric video, whereas naive 4DGS, which computes the temporal range for all 4D Gaussians, incurs large GPU memory usage with poor sampling efficiency for long video data.
Moreover, there is no speed downgrade for rendering because this memory copying process can be parallelized with the rendering.

Because the segment lengths of adjacent levels have an integer ratio, segments at different levels often have the same starting time.
This could result in increased usage of GPU memory during rendering.
In this case, many 4D Gaussians with short influence radius can span two adjacent segments across many levels.
According to \cref{eq:gaussian_node_assignment}, this tends to make the Gaussian appear in the longer segments, which in turn raises its probability of being sampled for rendering at many timestamps, leading to an increase in GPU memory usage.

To solve this issue, we propose a corrected offset $\overline{\tau}_{l}$ of level $l$ by half of the segment length of the next level at different levels to avoid these situations:
\begin{equation}
    \overline{\tau}_{l} = -\frac{S}{2^{l+2}},
\end{equation}
With this corrected offset, the start and end timestamp of \cref{eq:node_influence_range} and the index $n^t$  of \cref{eq:efficient_sampling} can be rewritten as:
\begin{align}
     & \overline{\tau}_{l,n} = \overline{\tau}_{l} + n \cdot s_{l},  \textit{\ \ }
    \underline{\tau}_{l,n} = \overline{\tau}_{l} + (n+1) \cdot s_{l},                                           \\
     & n^{t}_{l} = \mathrm{floor}(\frac{t - \overline{\tau}_{l}}{s_{l}}).     \label{eq:efficient_sampling_new}
\end{align}

\subsubsection{Hierarchy Update}
After each training step, the properties of the streaming Gaussians $\mathcal{G}_t$ are optimized with the reconstruction loss, resulting in an updated set $\mathcal{G}_t^\prime$ with varying influence radius as reflected by \cref{eq:gaussian_influence_range}.
So it is necessary to reassign the optimized 4D Gaussians back into their respective levels and nodes in the hierarchical structure using \cref{eq:gaussian_node_assignment}.

Specifically, for the updated Gaussian with a new opacity distribution $o^{\prime}_t \camera{\sim} \mathcal{N}(t;\mu^{\prime}_t,\sigma^{\prime}_t)$, the influence range $[\overline{\tau}^{\prime},\underline{\tau}^{\prime}]$ can be computed using \cref{eq:gaussian_influence_range}, with the same opacity threshold $o_{th}$.
Subsequently, the corresponding indices $\overline{n}^{\prime}_{l}, \underline{n}^{\prime}_{l}$ for the start and end influence timestamps $\overline{\tau}^{\prime},\underline{\tau}^{\prime}$ are computed for each level of the Gaussian hierarchy using \cref{eq:efficient_sampling_new}.
We can determine whether the Gaussian can adapt to this level $l$ by comparing whether $\overline{n}^{\prime}_{l}$ equals $\underline{n}^{\prime}_{l}$.
Then, the updated level $l^{\prime}$ is defined to be the maximum level satisfying this condition.
Note that since our update algorithm operates on levels instead of individual segments, the computational complexity remains $O(L)$, where $L$ is the total number of levels in the hierarchy as defined in \cref{sec:temporal_gaussian_hierarchy}.
Since $L$ is a constant with respect to the length of the represented volumetric video, this helps maintain the constant memory usage and iteration speed of our method.

\subsection{\cpm}
\label{sec:compact_appearance}

Original 3DGS~\cite{kerbl20233d} utilizes standard spherical harmonics (SH) evaluation, denoted as $\mathrm{evalSH}(\cdot)$, to obtain the view-dependent color $\mathbf{c} = \mathbf{c}_{base} + \mathrm{evalSH}(\mathbf{h}, \mathbf{d})$, where $\mathbf{h} \in \mathrm{R}^{m}$ represents the residual SH coefficients with $m$ denoting the number of SH bases, $\mathbf{d}$ denotes the view direction and $\mathbf{c}_{base}$ is the 3-channel base color of the Gaussian, \camera{resulting in a total of $3 \times (m + 1) ^ 2$ parameters for appearance per Gaussian}.
However, the large number of bases $m$ in standard SH representation results in significant storage overhead, particularly when scaling to long videos.

To address this issue, we propose a \cpm that leverages sparse SH to efficiently capture view-dependent effects with a reduced model size while maintaining comparable rendering quality to the full-SH model.
The diffuse part of the volumetric video does not necessarily require high-degree SH for accurate modeling, inspiring us to exclude this type of Gaussians to reduce the storage.
Intuitively, diffuse Gaussians should receive a low magnitude gradient on its residual SH coefficients $\mathbf{h}$ even when they are rendered and optimized as view-dependent ones.
Thus we can use a zero-initialization and gradient thresholding strategy to successfully identify these Gaussians and retain their diffuse properties.
Specifically, we first initialize the residual SH coefficients $\mathbf{h}$ of all Gaussians with zeros.
Then, a gradient threshold $g_{th}$ is defined to modify the gradient values of the low-magnitude ones $||\mathbf{g}_{\mathbf{h}}||_2 < g_{th}$ as zero.
For the view-dependent Gaussians, we optimize them regardless of their gradient scale to enable our model to express view-dependent effects.
This process can be described using:
\begin{equation}
    \mathbf{g}^{\prime}_{\mathbf{h}} = \left\{
    \begin{array}{ll}
        \mathbf{g}_{\mathbf{h}} , & ||\mathbf{g}_{\mathbf{h}}||_2 \geq g_{th} \lor ||\mathbf{h}||_2 \neq 0 \\
        0 ,                       & ||\mathbf{g}_{\mathbf{h}}||_2 < g_{th} \land ||\mathbf{h}||_2 = 0
    \end{array} \right..
\end{equation}
To regulate the number of view-dependent Gaussians, we additionally compute its proportion and set the gradient threshold $g_{th}$ to infinity for future optimization steps once the proportion reaches a certain ratio $\lambda_{\mathbf{h}}$.

After training, we group the 4D Gaussians according to whether they are diffuse or view-dependent.
This ensures that all zeros in the SH coefficients of all points $\mathcal{C}_d$ can be grouped together before being fed into the Huffman Coding algorithm \cite{huffman1952method}, further boosting the compression ratio.

\subsection{Training}
\label{sec:training}

Given the rendered image $\mathbf{I}$ and ground-truth image $\mathbf{I}_{gt}$, we optimize our model with the following objective:
\begin{equation}
    \mathcal{L} = \lambda_{m} \cdot \mathcal{L}_{mse} + \lambda_{s} \cdot \mathcal{L}_{ssim} + \lambda_{p} \cdot \mathcal{L}_{perc},
\end{equation}
where $\mathcal{L}_{mse}$ denotes the mean squared error using
\begin{equation}
    \mathcal{L}_{mse} = || \mathbf{I} - \mathbf{I}_{gt} || _ 2 ^ 2,
\end{equation}
$\mathcal{L}_{ssim}$ denotes the structural similarity \cite{wang2004image} error
\begin{equation}
    \mathcal{L}_{ssim} = 1 - \mathrm{SSIM}(\mathbf{I}, \mathbf{I}_{gt}),
\end{equation}
and $\mathcal{L}_{perc}$ denotes the perceptual similarity \cite{zhang2018unreasonable} loss by computing the L$_1$ difference between the extracted features with AlexNet \cite{krizhevsky2012imagenet} (denoted by $\mathrm{\Phi}$):
\begin{equation}
    \mathcal{L}_{perc} = || \mathrm{\Phi}(\mathbf{I}) - \mathrm{\Phi}(\mathbf{I}_{gt}) ||_{1}.
\end{equation}
$\lambda_{m}$, $\lambda_{s}$ and $\lambda_{p}$ are loss weights of $\mathcal{L}_{mse}$, $\mathcal{L}_{ssim}$ and $\mathcal{L}_{perc}$.
Following 3DGS \cite{kerbl20233d}, we also apply the adaptive control scheme to split, clone and prune the 4D Gaussian ellipsoids every 100 iterations using their strategy.
These newly split and cloned ellipsoids can be easily updated into \tgh using the algorithm described in \cref{sec:temporal_gaussian_hierarchy}.
More details can be found in \supp.

\subsection{Real-Time Rendering}
\label{sec:rendering}

To facilitate optimization of the Gaussians, 3DGS \cite{kerbl20233d} developed a CUDA-based software rasterizer for their Gaussian Splatting algorithm.
To further improve the rendering speed during inference, we develop a new hardware-accelerated rasterization algorithm \camera{that combines CUDA-based sorting and Graphics-based rasterization} to replace the standard software rasterizer.

Given the timestamp $t$, we acquire the streaming Gaussians set $\mathcal{G}_{t}$ using \cref{eq:efficient_sampling} and arrange them in back-to-front order based on view-space depth with a fast radix sort on the GPU \cite{paszke2019pytorch}.
These sorted Gaussians are then transferred to the hardware rasterizer \cite{shreiner2009opengl} and projected into screen space using the 3D Gaussian Splatting algorithm, resulting in 2D Gaussians.
Since 2D Gaussians are not directly recognized by the rasterizer for computing gaussian-pixel pairs, we transform the 2D Gaussian into a 2D rectangular primitive (quad) with opacity thresholding \camera{\cite{mkkellogg2024GaussianSplats3D,3dgstutorial}}.
The hardware rasterization pipeline \cite{shreiner2009opengl} then rasterizes the set of quads onto the screen, generating quad-pixel pairs (fragments) in which we can easily compute the opacity and color for the pixel locations by evaluating the multivariant Gaussian distribution function.
Since the Gaussians are sorted in back-to-front order, the hardware rasterizer can efficiently perform back-to-front alpha blending \cite{mildenhall2021nerf} on all fragments corresponding to a particular pixel to compute its final color, which is concatenated to be the final rendered image.
By directly utilizing the hardware rasterization pipeline and using the GPU for sorting, this implementation can lead to better utilization of the GPU and lead to a significant improvement in rendering efficiency.
As demonstrated in \cref{fig:ablation-components}, our approach achieves a 5x speed improvement over the standard software rasterization process \cite{kerbl20233d}.
\camera{The implementation is open-sourced at \href{https://github.com/dendenxu/fast-gaussian-rasterization}{https://github.com/dendenxu/fast-gaussian-rasterization}.}

\begin{figure*}
    \includegraphics[width=1.0\linewidth]{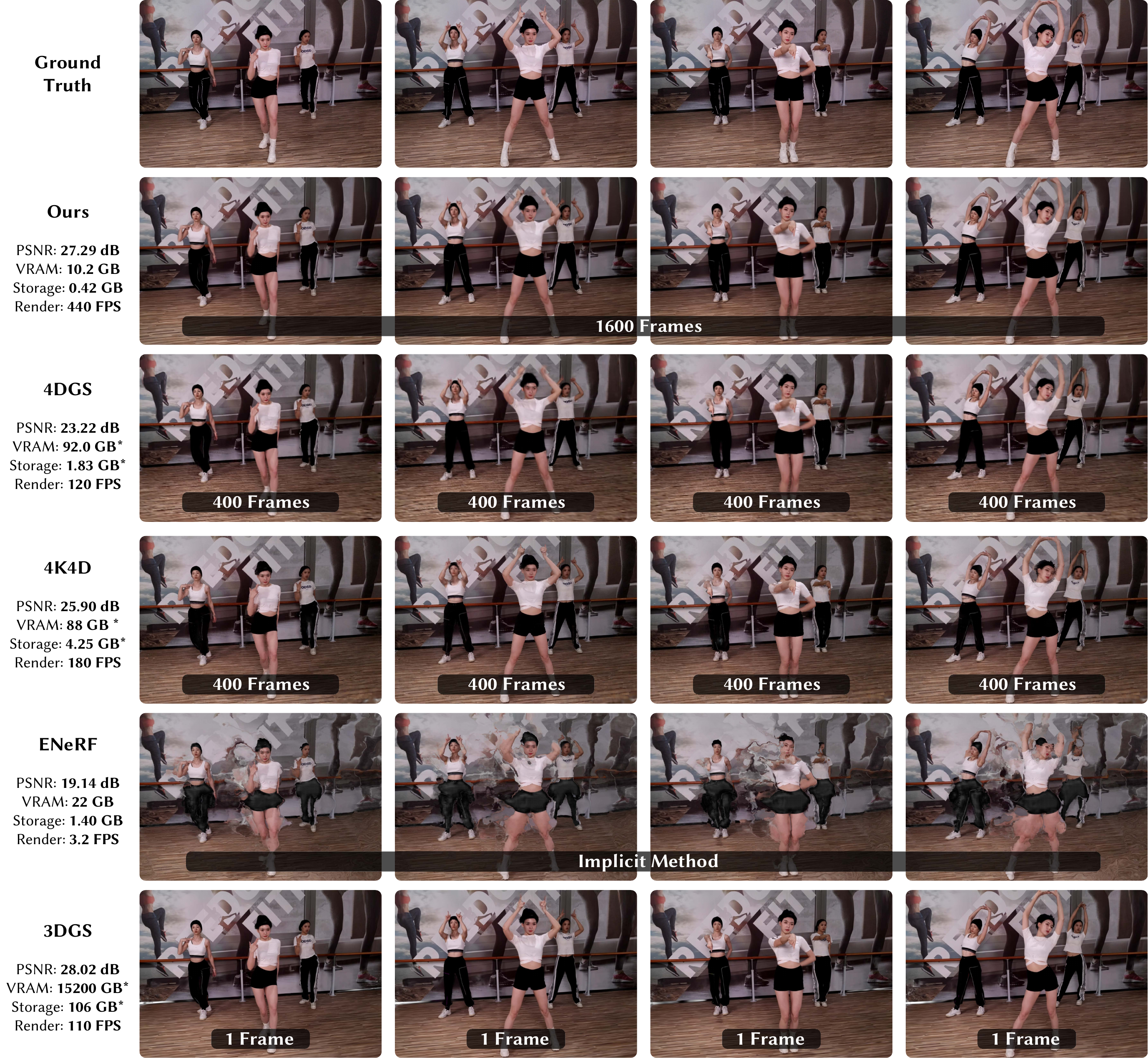}
    \caption{
        \textbf{Qualitative comparisons on \textit{Mobile-Stage}~\cite{xu20244k4d} with 1600 frames.} For long videos of  1200 frames, our model can be directly trained on the whole sequence and only requires 10.2GB of VRAM for training and 0.42GB of storage, which is 2x and 4x less, respectively, compared to the second smallest implicit method, ENeRF~\cite{lin2022efficient}.
        Contrarily, 4K4D \cite{xu20244k4d} and 4DGS \cite{yang2023realtime} could only be trained on small segments of 300 frames without encountering Out-of-Memory error.
        Our model achieves high rendering quality and can be rendered at 440 FPS.
    }
    \label{fig:comparison_ms}
\end{figure*}

\begin{figure*}
    \includegraphics[width=1.0\linewidth]{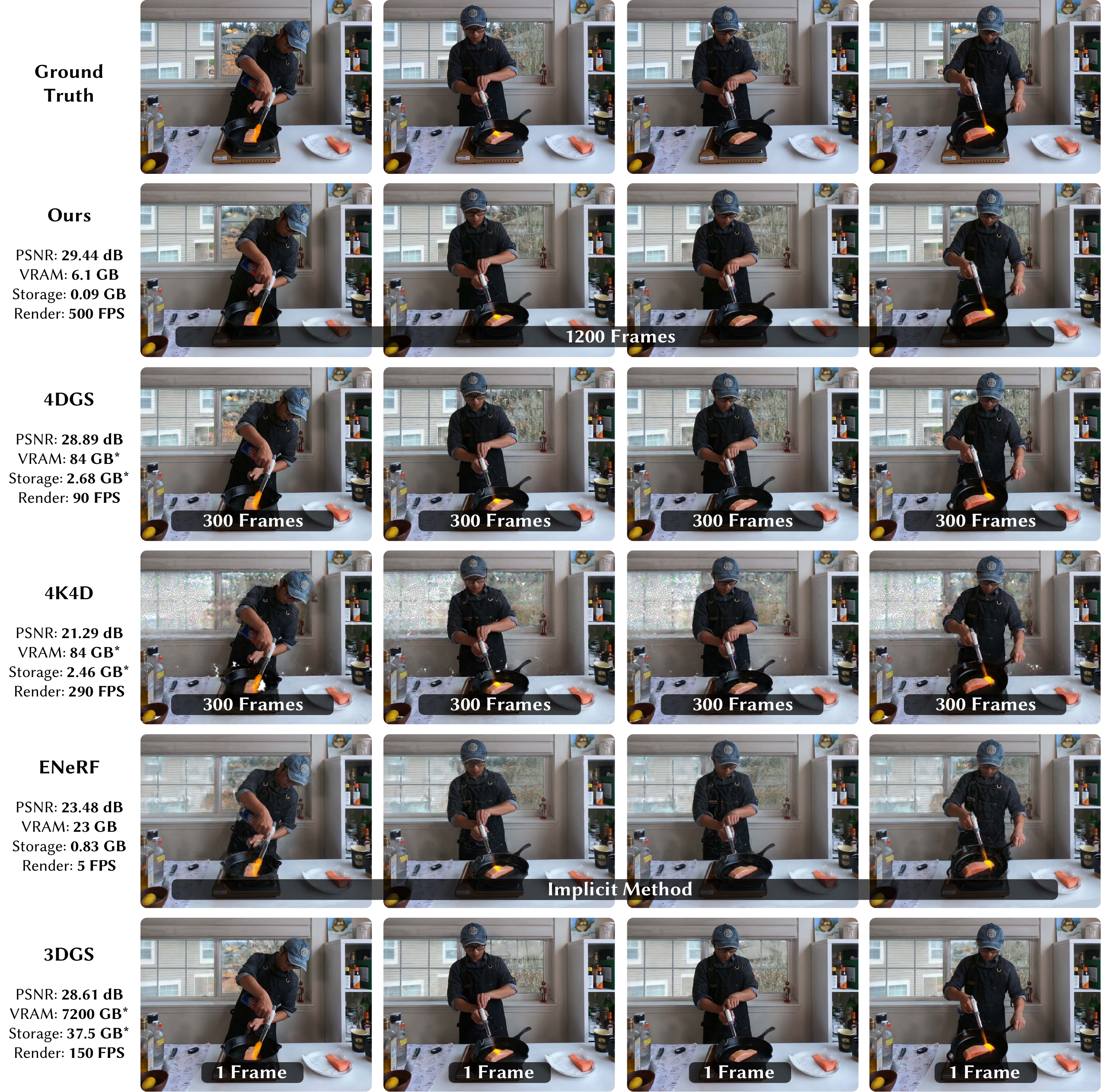}
    \caption{\textbf{Qualitative comparisons on \textit{Neural3DV}~\cite{li2022neural} with 1200 frames.} Our method can not only recover high-frequency details of dynamic objects but also maintain the sharp appearance of the background with low training costs and a compact model size.
        4K4D \cite{xu20244k4d} and 4DGS \cite{yang2023realtime} could only be trained on small segments of 300 frames due to VRAM limitation.
    }
    \label{fig:comparison_n3dv}
\end{figure*}

\begin{figure*}
    \includegraphics[width=1.0\linewidth]{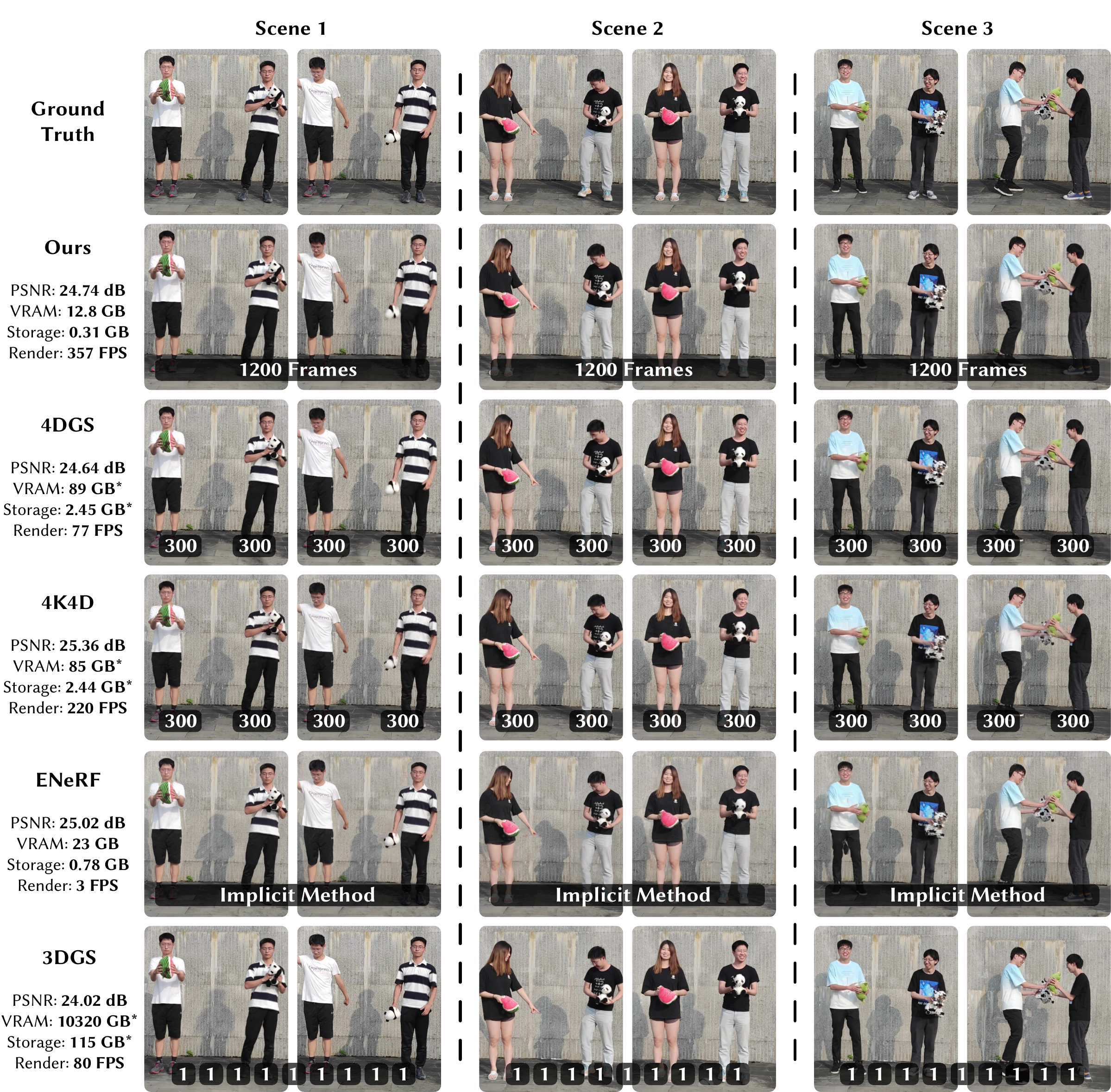}
    \caption{\textbf{Qualitative comparisons on \textit{ENeRF-Outdoor}~\cite{lin2022efficient} with 1200 frames.}
        Here we show multiple sequences for comparison. Our method achieves high-quality rendering while using only 50\% of the VRAM and 40\% of the storage required by ENeRF, and it is 1.6x faster than the second best method, 4K4D.
        Note that 4K4D \cite{xu20244k4d} and 4DGS \cite{yang2023realtime} encounters Out-of-Memory error for sequences longer than 300 frames.
    }
    \label{fig:comparison_eo}
\end{figure*}

\section{Implementation Details}
\label{sec:implementation_details}

\paragraph{Optimization}
Our \method representation is trained using the PyTorch framework \cite{paszke2019pytorch,xu2023easyvolcap} along with custom CUDA \cite{sanders2010cuda} kernels for efficient update of the \tgh structure.
We optimize the model using an Adam \cite{kingma2014adam} optimizer with a learning rate of $1.6e^{-4}$ following 3DGS \cite{kerbl20233d}.
Typically, a model is trained for 50k iterations for a sequence length of 1200 frames, which takes around 2 hours on an RTX 4090 GPU.
We scale the number of iterations linearly with the sequence length.
Notably, different datasets come in different resolutions, thus the speed for an iteration might vary accordingly, leading to varying training times.
Following 3DGS \cite{kerbl20233d} and 4DGS\cite{yang2023realtime}, we perform adaptive control on the number of Gaussians by splitting, cloning, and pruning the Gaussians based on their view-space gradients and opacity values every 100 iterations.
Thanks to our efficient \tgh structure, this adaptive control can be limited to only the sampled segments during the previous iterations, which significantly lower the GPU memory usage and computational cost for this stage.

\paragraph{Hyperparameters}
Hyperparameters of the \tgh structure and the Compact Appearance model are set empirically according to the sensitivity analysis in \cref{sec:ablation} or directly followed from previous work.
Specifically, the number of levels $L$ in the \tgh structure is set to 9 and the root segment $S$ size is set to $10s$ throughout evaluation.
Following 4DGS \cite{yang2023realtime}, the opacity threshold for the marginal temporal Gaussian $o_{th}$ is set to $0.05$.
For the Compact Appearance model, we set the gradient threshold $g_{th}$ for separating diffuse and view-dependent Gaussians to $1.0e^{-6}$ and the cut-off ratio $\lambda_{h}$ to $0.15$.
Similar to 3DGS \cite{kerbl20233d}, we use 3 degrees of Spherical Harmonics for the view-dependency modeling, resulting in $m = 48$ SH basis functions.
The weights for the mean squared error and structure and perceptual similarity loss $\lambda_{m}, \lambda_{s}, \lambda_{p}$ are empirically set to 0.8, 0.2, and 0.01, respectively.

\paragraph{Initialization}
Following 3DGS \cite{kerbl20233d}, we use the sparse point clouds output by the SfM process \cite{schonberger2016structure} for every frame of the volumetric video for initialization.
For methods utilizing a different initialization strategy \cite{yang2023realtime,xu20244k4d}, we adapt their initial number of points to be at least as large as ours by upsampling their initial point clouds to ensure a fair comparison.
The initial 3D scaling $\mathbf{s}_{x,y,z}$ of the Gaussians is determined by performing a KNN sampling on the sparse point cloud and computing the average distances following 3DGS \cite{kerbl20233d}.
The temporal scaling is $\mathbf{s}_{t}$ is set to frame the time of the sequence to be reconstructed.

\section{Experiments}

\newcommand{\cmr}{ss}
\begin{table*}[ht]
    \begin{center}
        \caption{
            \textbf{Quantitative comparison on \textit{Neural3DV}~\cite{li2022neural}, \textit{ENeRF-Outdoor}~\cite{lin2022efficient}, \textit{MobileStage}~\cite{xu20244k4d} \camera{and \textit{CMU-Panoptic}~\cite{joo2015panoptic}}}. We report PSNR, SSIM, and LPIPS to evaluate rendering quality. VRAM and storage are used to assess the training and storage costs. "Train." and "Render." denote the training time and rendering FPS, respectively. Note that we highlight the best one for clarification. 4K4D, 4DGS and 3DGS could only be trained in 300-frame or 1-frame segments. "*" indicates that the computational costs for these methods are summed over the entire video. \camera{"**" indicates that the method uses the same Gaussian initialization as ours.}
        }
        \label{tab:main_comparison}
        \resizebox{1.0\textwidth}{!}{
            \setlength\tabcolsep{1.0pt}
            \begin{tabular}{l|cccccc|ccccc|cccccc|cc}
                \toprule
                \multirow{2}{*}{\makecell[c]{Metrics}} & \multicolumn{6}{c|}{Neural3DV \cite{li2022neural}} & \multicolumn{5}{c|}{ENeRF-Outdoor \cite{lin2022efficient}} & \multicolumn{6}{c|}{MobileStage \cite{xu20244k4d}} & \multicolumn{2}{c}{\camera{CMU-Panoptic}}                                                                                                                                                                                                                                                 \\
                                                       & Ours                                               & 4DGS                                                       & 4K4D                                               & ENeRF                                     & 3DGS     & \camera{Dy3DGS          } & Ours             & 4DGS     & 4K4D           & ENeRF   & 3DGS      & Ours             & 4DGS     & \camera{4DGS** } & 4K4D     & ENeRF          & 3DGS            & \camera{Ours   }          & \camera{Dy3DGS } \\
                \midrule
                PSNR  $^{\uparrow}$                    & \textbf{29.44}                                     & 28.89                                                      & 21.29                                              & 23.48                                     & 28.61    & \camera{25.91           } & 24.74            & 24.64    & \textbf{25.36} & 25.02   & 24.02     & 27.29            & 23.21    & \camera{24.03  }  & 25.90    & 19.14          & \textbf{28.02}  & \camera{\textbf{28.55  }} & \camera{24.27  } \\ %
                SSIM $^{\uparrow}$                     & 0.9450                                             & \textbf{0.9521}                                            & 0.8266                                             & 0.8944                                    & 0.9498   & \camera{0.8809          } & \textbf{0.8392}  & 0.7855   & 0.8080         & 0.7824  & 0.8231    & 0.9127           & 0.7876   & \camera{0.8150 }  & 0.8788   & 0.7492         & \textbf{0.9172} & \camera{\textbf{0.9558 }} & \camera{0.9432 } \\ %
                LPIPS $^{\downarrow}$                  & 0.2144                                             & \textbf{0.1968}                                            & 0.3715                                             & 0.2599                                    & 0.2103   & \camera{0.2555          } & \textbf{0.2624}  & 0.3118   & 0.3795         & 0.3043  & 0.2765    & 0.2536           & 0.4209   & \camera{0.3880 }  & 0.3872   & 0.4365         & \textbf{0.2383} & \camera{\textbf{0.4016 }} & \camera{0.5135 } \\ %
                \midrule
                VRAM $^{\downarrow}$                   & 6.1 GB                                             & 84 GB*                                                     & 84 GB*                                             & 23 GB                                     & 7200 GB* & \camera{\textbf{5.0 GB} } & \textbf{12.8 GB} & 89 GB*   & 85 GB*         & 23 GB   & 10320 GB* & \textbf{10.2 GB} & 92 GB*   & \camera{88 GB*}  & 88 GB*   & 22 GB          & 15200 GB*       & \camera{\textbf{10.0 GB}} & \camera{11.4 GB} \\ %
                Storage $^{\downarrow}$                & \textbf{0.09 GB}                                   & 2.68 GB*                                                   & 2.46 GB*                                           & 0.83 GB                                   & 37.5 GB* & \camera{19.5 GB         } & \textbf{0.31 GB} & 2.45 GB* & 2.44 GB*       & 0.78 GB & 115 GB*   & \textbf{0.42 GB} & 1.83 GB* & \camera{2.51 GB*}  & 4.25 GB* & 1.40 GB        & 106 GB*         & \camera{\textbf{0.22 GB}} & \camera{6.42 GB} \\ %
                Train. $^{\downarrow}$                 & \textbf{2.1 h}                                     & 10.4 h*                                                    & 26.6 h*                                            & 4.6 h                                     & 110 h*   & \camera{37.1 h          } & \textbf{4.4 h}   & 12.7 h*  & 29.3 h*        & 6.5 h   & 280 h*    & 10.4 h           & 14.4 h*  & \camera{12.7 h* }  & 40.5 h*  & \textbf{6.1 h} & 213 h*          & \camera{\textbf{10.8 h }} & \camera{20.8 h } \\ %
                Render. $^{\uparrow}$                  & 550 FPS                                            & 90 FPS                                                     & 290 FPS                                            & 5 FPS                                     & 150 FPS  & \camera{\textbf{610 FPS}} & \textbf{357 FPS} & 77 FPS   & 220 FPS        & 3 FPS   & 80 FPS    & \textbf{440 FPS} & 120 FPS  & \camera{86 FPS}  & 180 FPS  & 3 FPS          & 110 FPS         & \camera{\textbf{475 FPS}} & \camera{305 FPS} \\

                \bottomrule
            \end{tabular}
        }
    \end{center}
\end{table*}

\subsection{Settings}

\subsubsection{Datasets}
\label{sec:datasets}
We use public datasets \textit{Neural3DV}~\cite{li2022neural}, \textit{ENeRF-Outdoor}~\cite{lin2022efficient}, \textit{MobileStage}~\cite{xu20244k4d,xu2024relightable} \camera{and \textit{CMU-Panoptic} \cite{joo2015panoptic}} for evaluation, since they provide longer video sequences. \textit{Neural3DV} is a popular benchmark for novel view synthesis. It is captured by a multi-view system with 19-21 cameras. The videos are recorded at a resolution of 2704$\times$2028 and 30FPS. We select the \textit{flame\_salmon} scene for evaluation, which uses 19 cameras and contains 1200 frames in total. In line with prior work, we downsample the videos by a factor of two and follow the training and testing camera split as specified by~\cite{li2022neural}.
\textit{ENeRF-Outdoor} is a dynamic dataset of outdoor scenes, collected by 18 synchronized cameras at 1920$\times$1080 and 60FPS. We select three 1200-frame sequences for evaluation: \textit{actor1\_4}, \textit{actor2\_3} and \textit{actor5\_6}. Each sequence contains two actors holding objects in an outdoor environment. We use the `08' camera as the testing view and the remaining views as the training views.
\textit{MobileStage} is a multi-view dataset focused on dynamic humans. The dataset records actors at 30FPS with 24 1080p cameras. We evaluate on the \textit{dance3} sequence, which captures 3 dancing actors for 1600 frames. The `05' camera is used for testing and the remaining cameras are for training. The dataset is challenging for dynamic view synthesis due to the complex motion and fast movement involved in dancing.
\camera{
    \textit{CMU-Panoptic} \cite{joo2015panoptic} is a large-scale multi-view dataset capturing varing everyday human interactions and activities. We follow Dy3DGS \cite{luiten2023dynamic} to select three subsequences from the \textit{sports} clip, namely the \textit{box}, \textit{softball} and \textit{basketball} subsequences, and also follow their training and testing split of 27 to 4. Contrary to Dy3DGS \cite{luiten2023dynamic}, we use the full resolution of the 31 HD cameras and the full length of the clip, resulting in a resolution of 1080p and a length of 1000-frame, 800-frame and 700-frame respectively for the three subsequences.
}

In addition to leveraging existing public datasets, we build a new dynamic multi-view dataset named \textit{SelfCap} to evaluate our approach.
This dataset comprises three dynamic videos, each captured at 60FPS in 4K resolution using a synchronized array of 22 iPhone cameras.
The videos in this dataset range from 2 to 10 minutes in length, significantly surpassing the durations of previous datasets.
By training and evaluating our methods on these long video sequences, we demonstrate our ability to efficiently represent long volumetric videos.
We will publicly release this dataset to facilitate research on long volumetric video modeling.

\camera{All datasets are captured using synchronized static camera arrays, and no explicit temporal consistency is enforced other than sharing the camera parameters.}
\camera{The datasets mainly contain static backgrounds with dynamic foregrounds for everyday dynamic activities involving dynamic humans and objects that are mostly diffuse, which further justifies the use of the global segment and Compact Appearance model.}
\camera{Noticeably, our representation is defined on the world coordinate system, thus even if the cameras are moving, it should not have too much impact as long as the camera parameters are known.}

\subsubsection{Baselines}
We include several state-of-the-art baseline methods for comparison, including 3DGS~\cite{kerbl20233d}, 4DGS~\cite{yang2023realtime}, ENeRF ~\cite{lin2022efficient}, 4K4D~\cite{xu20244k4d} \camera{and Dy3DGS~\cite{luiten2023dynamic}}.
3DGS employs explicit Gaussian primitives to effectively model 3D scenes.
While this technique achieves high-quality rendering and impressive speed, its application is limited to static scenes.
To extend its evaluation to dynamic scenes, we experiment with training 3DGS on a per-frame basis for the testing frames for comparison.
\camera{
    Dy3DGS \cite{luiten2023dynamic} optimizes a per-frame tracking of the scene geometry and appearance using 3DGS \cite{kerbl20233d} in a streaming fashion.
}
4DGS expands 3DGS along the temporal axis and introduces a 4D representation to capture the spatial-temporal dynamics.
By leveraging 4D primitives that encompass both spatial and temporal dimensions, it enables explicit geometry and appearance modeling in volumetric videos.
Besides the explicit approaches, we also consider dynamic approaches that utilize implicit and hybrid representations.
ENeRF constructs a cascade cost volume to predict coarse scene geometry, which is then used to guide point sampling for generative implicit radiance fields.
4K4D integrates dynamic point clouds as input and designs a hybrid appearance model to improve rendering quality.
Baseline models are trained with the official implementation at different frame-length settings. For 3DGS, we train it in a per-frame fashion. For 4DGS and 4K4D, we split each video into 4 equal-length segments, and trained them on each segment individually. As for ENeRF, we finetune the official pre-trained model on the whole video.

\subsubsection{Metrics}
We employ the PSNR, SSIM~\cite{ssim}, and LPIPS~\cite{lpips} metrics to assess the quality of rendered images in our methods and baselines.
\textit{PSNR} measures the $l_2$ difference between a reconstructed image and its ground truth, with higher values indicating less disparity.
\textit{SSIM} quantifies the structural similarity between images, with values ranging from -1 to 1, where higher values denote greater similarity.
\textit{LPIPS} evaluates perceptual similarity, aligned with human perception.

Besides, we consider VRAM, storage, training time, and rendering speed to illustrate training and inference costs.
\textit{VRAM} represents the GPU memory required for training, measured in gigabytes (GB).
\textit{Storage} indicates the model size on disk, measured in gigabytes (GB).
Baseline methods such as 3DGS, 4K4D, and 4DGS are incapable of handling extensive volumetric videos. For comparison, we train these models on short segments and combine them into the final model. When the video is divided into $M$ segments, VRAM and storage requirements scale $M$ times accordingly.
Moreover, since 4K4D and ENeRF require the source images as input, their reported storage cost also includes the video-encoded images.
\textit{Training time} reflects the convergence speed of each method. Similar to the \textit{storage} cost, the training cost is summed over all segments for 3DGS, 4K4D, and 4DGS.
\textit{Rendering speed} is evaluated as the rendering FPS of each method when rendering to a real-time GUI.

\begin{figure*}[h]
    \includegraphics[width=1\linewidth]{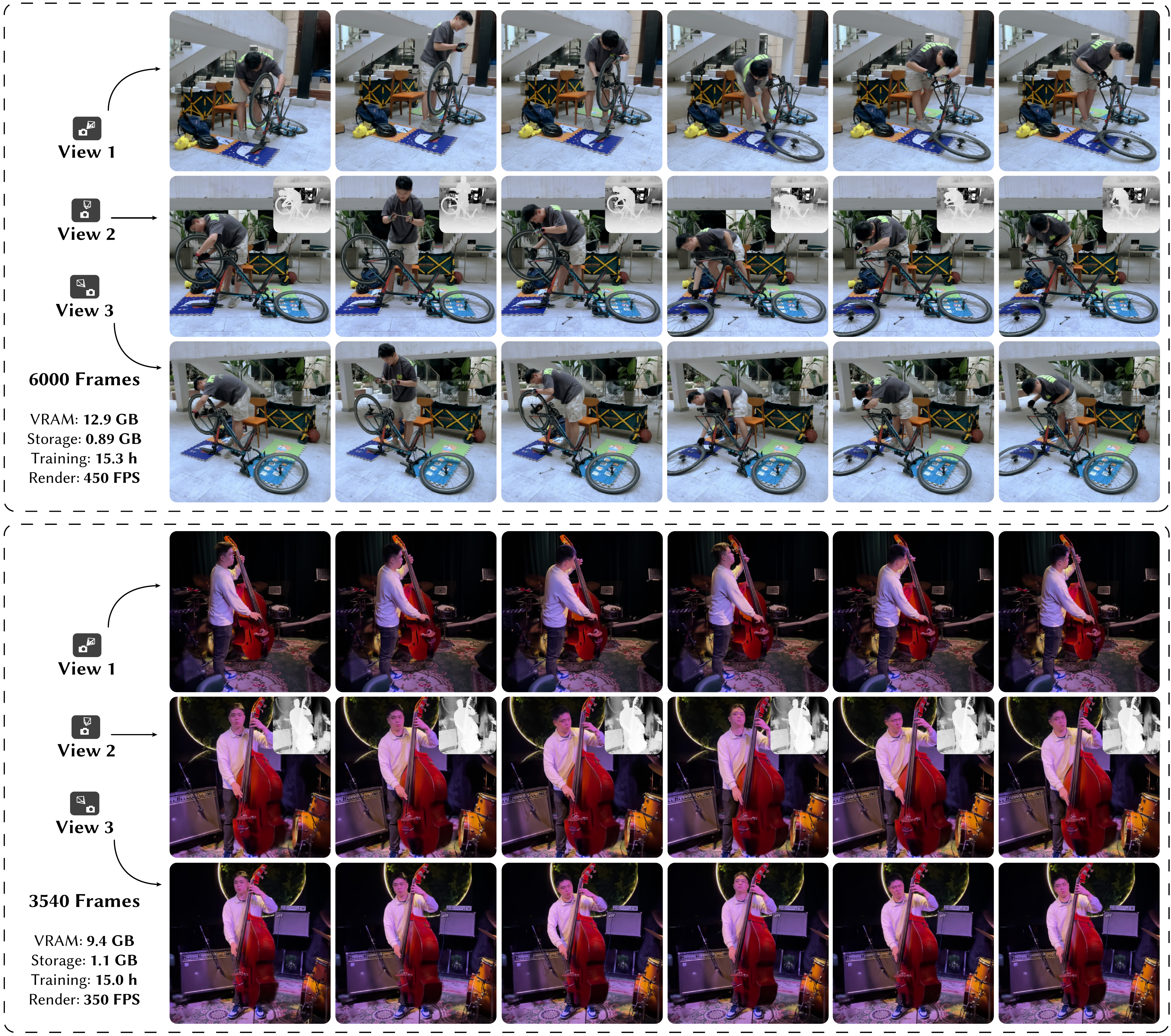}
    \caption{
        \textbf{Qualitative results on long videos from our \textit{SelfCap}.}
        We evaluate our method on very long videos consisting of 6000 and 3540 frames.
        Previous methods either suffer from high computational costs for training or low rendering quality, while our method addresses these challenges.
        \camera{The total number of 4D Gaussian points for these long sequences are 9.83M and 15.06M respectively.}
    }
    \label{fig:long_video}
\end{figure*}

\subsection{Comparison Experiments}

Both qualitative and quantitative comparisons are presented in \cref{tab:main_comparison} and \cref{fig:comparison_n3dv,fig:comparison_eo,fig:comparison_ms}.
As shown in \cref{tab:main_comparison}, our method achieves the best or comparable overall performance on all datasets across all evaluation metrics, including training cost, memory usage, speed, and image quality.

Compared to Gaussian-based methods like 3DGS and 4DGS, our model benefits from lower training costs, storage requirements, and faster inference speeds.
While 3DGS~\cite{kerbl20233d} achieves good visual quality through per-frame training, its VRAM and storage usage scales unfavorably with the length of the video, making it unsuitable for handling long volumetric videos.
Additionally, since 3DGS models each frame independently, it often produces videos with noticeable flickering artifacts.
4DGS~\cite{yang2023realtime} extends 3DGS with a temporal axis to improve its capacity to model temporal dynamics but still suffers from high computation costs for training and storage.

Compared with the implicit method ENeRF~\cite{lin2022efficient}, our model achieves an 80x speed improvement because ENeRF requires costly network evaluations for appearance modeling.
As shown in Fig. \ref{fig:comparison_n3dv}, ENeRF tends to produce blurry results around occlusions and edges on the datasets with complex geometry like \textit{MobileStage}~\cite{xu20244k4d}, degrading the visual quality a lot.

As for the hybrid method 4K4D~\cite{xu20244k4d}, it maintains a comparable inference speed to ours. However, the use of point clouds as geometry representation limits its ability to provide highly detailed appearances, especially on \textit{Neural3DV}~\cite{li2022neural}, where point clouds are challenging to estimate accurately due to complex backgrounds.
Moreover, like 3DGS, 4K4D employs a per-frame modeling approach, resulting in flickering artifacts.

\begin{figure*}[t]
    \includegraphics[width=1.0\linewidth]{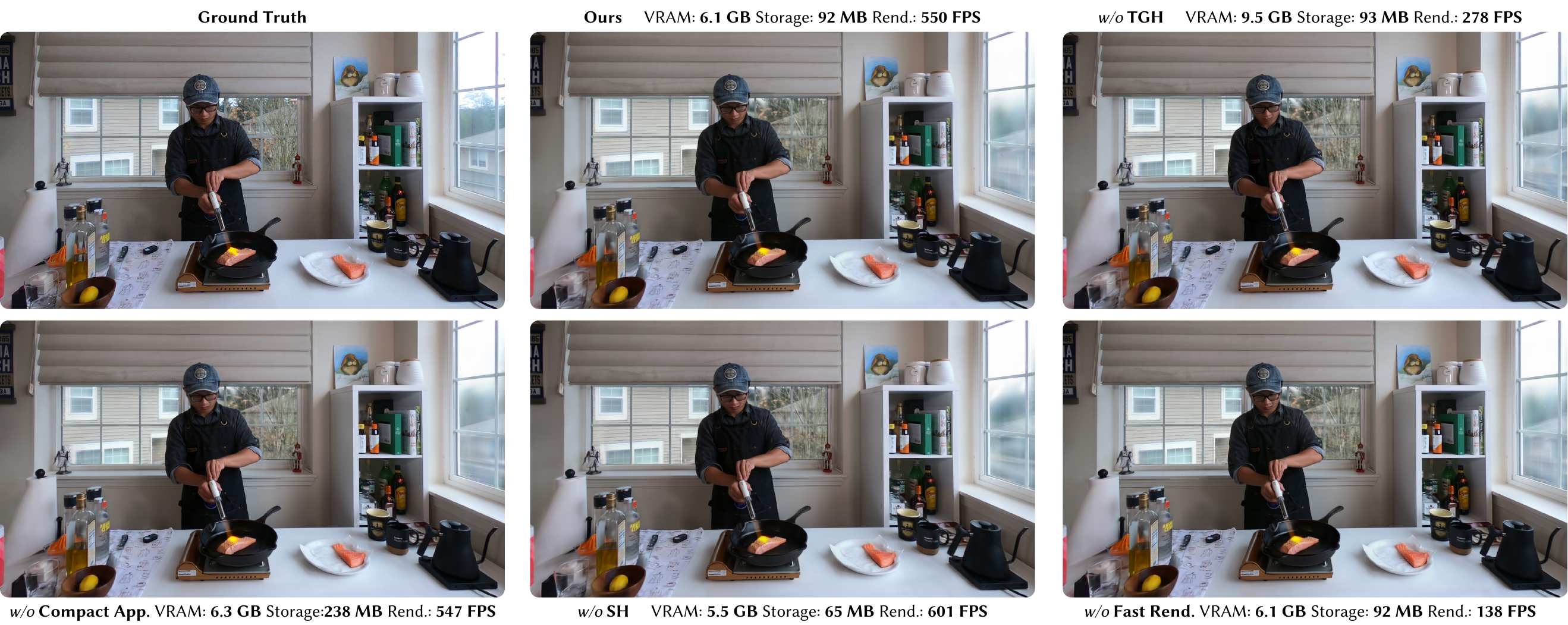}
    \caption{\textbf{Qualitative comparison for ablations on proposed components.}}
    \label{fig:ablation-components}
\end{figure*}

\begin{table}[t]
  \caption{\textbf{Ablations on proposed components on the \textit{Neural3DV} \cite{li2022neural} dataset.}
    Our \tgh representation successfully reduces the GPU memory usage to a constant level and the Compact Appearance model significantly lowers our model size.
  }
  \label{tab:ablation-components}
  \resizebox{\linewidth}{!}{%
    \begin{tabular}{l|ccc|cccc}
      \toprule
                         & PSNR ${\uparrow}$ & LPIPS ${\downarrow}$ & SSIM ${\uparrow}$ & VRAM ${\downarrow}$ & Storage ${\downarrow}$ & Train.${\downarrow}$ & Render.${\uparrow}$ \\
      \midrule
      \textit{w/o} TGH   & 29.18             & 0.214                & 0.952             & 9.5 GB              & 0.093 GB               & 2.6 h                & 278 FPS             \\
      \textit{w/o} App.  & 29.01             & 0.214                & 0.953             & 6.3 GB              & 0.238 GB               & 2.3 h                & 547 FPS             \\
      \textit{w/o} Rend. & 29.17             & \textbf{0.213}       & 0.952             & 6.1 GB              & 0.092 GB               & 2.2 h                & 138 FPS             \\
      \midrule
      Ours               & \textbf{29.43}    & 0.214                & \textbf{0.954}    & \textbf{6.1 GB}     & \textbf{0. 92 GB}      & \textbf{2.2 h}       & \textbf{550 FPS}    \\
      \bottomrule
    \end{tabular}
  }
\end{table}

\subsection{Results on Long Videos of \textit{SelfCap}}
The aforementioned experiments demonstrate the feasibility of our method for videos ranging from 40$\sim$50s.
Compared to previous methods capable of handling only 1$\sim$2s, our approach achieves significant improvements.
To further validate the potential for longer videos, we test our method on the new dataset \textit{SelfCap}, which spans 6000 frames or even longer.
Benefiting from our \tgh, our method can maintain very low training costs, enabling it to represent extensive volumetric videos.
In \cref{fig:long_video}, we present visual results of videos from the \textit{SelfCap} dataset.
Our method showcases high-fidelity reconstruction results, particularly in complex areas with edges or occlusions, demonstrating superior rendering quality.

\subsection{Ablation Study}
\label{sec:ablation}
We analyze our model components and components design choices on the \textit{Neural3DV} \cite{li2022neural} dataset.

\subsubsection{Ablations on Model Components}
\ \\
The qualitative and quantitative results for ablations on model components are shown in \cref{fig:ablation-components} and \cref{tab:ablation-components}.
\ \\
\paragraph{\tgh}
We conduct experiments to evaluate the effectiveness of \tgh.
Without \tgh (first row of Tab. \ref{tab:ablation-components}), the model retrieves all 4D Gaussians to represent the training frames at each iteration.
However, a significant portion of these Gaussians may not be relevant, resulting in high VRAM usage and slow iteration speed.
Due to the substantial computational cost, the training is not performed sufficiently, resulting in degraded rendering results both quantitatively and qualitatively.

\paragraph{\cpm}
We investigate the effectiveness of \cpm by comparing the results in the 2nd and 4th rows of Tab. \ref{tab:ablation-components}.
Our model occupies only 40\% of the disk storage compared to the model with full spherical harmonics (SH) coefficients while maintaining the same rendering quality.
This demonstrates the potential for our model to be extended to longer volumetric videos without incurring large storage overhead.

\paragraph{Real-time Rendering}
We conducted an ablation study on the real-time rendering algorithm used in the inference stage.
Compared to the vanilla rasterization pipeline of 3DGS (3rd row, Tab. \ref{tab:ablation-components}), our algorithm achieves a 5x improvement in rendering speed, benefiting from the fast hardware rasterization.

\begin{figure*}
    \includegraphics[width=1\linewidth]{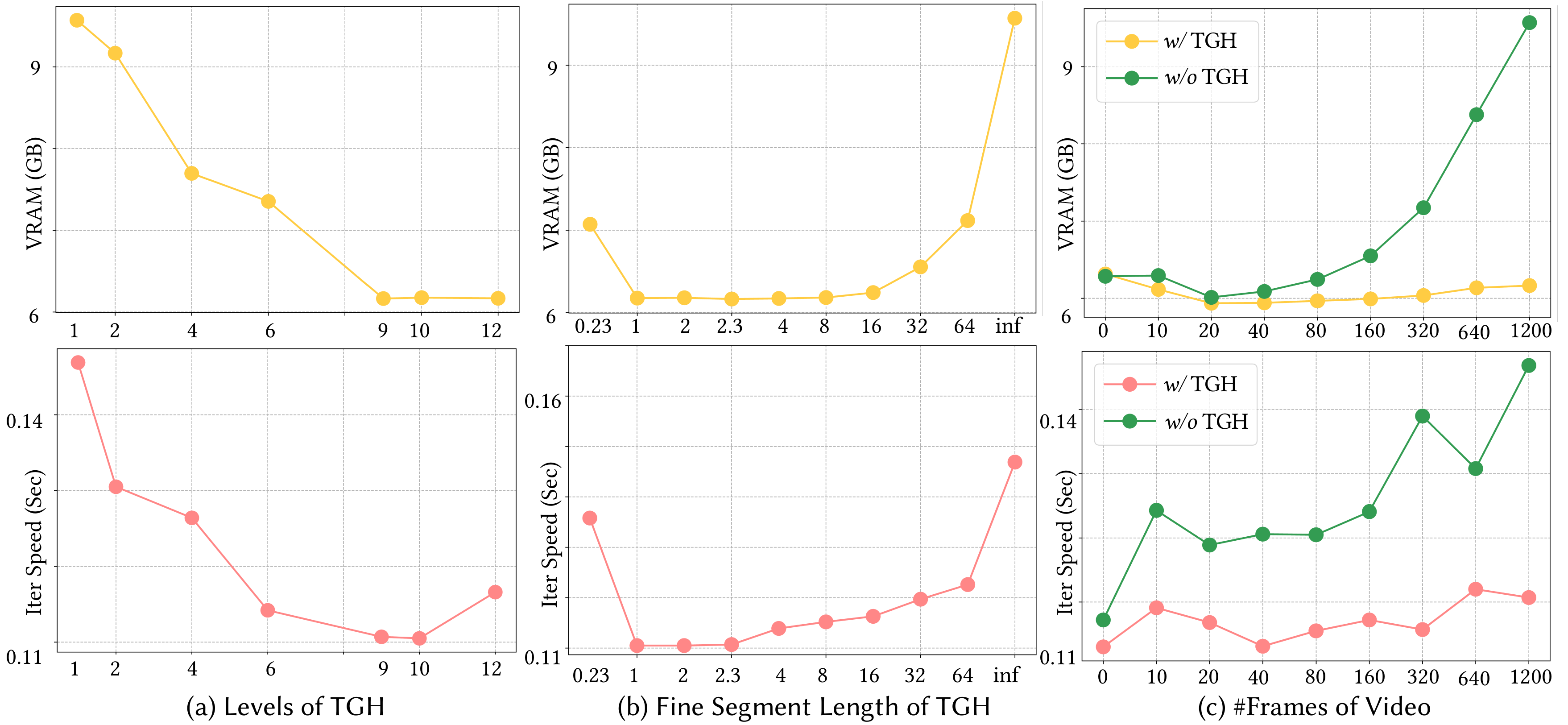}
    \caption{
        \textbf{Ablations on \tgh}:  We analyze the effects of level $L$, root segment length $S$, and the number of frames in a training video on computational costs, including VRAM usage and iteration speed.
    }
    \label{fig:ablation-tgh}
\end{figure*}

\begin{figure*}
    \includegraphics[width=1.0\linewidth]{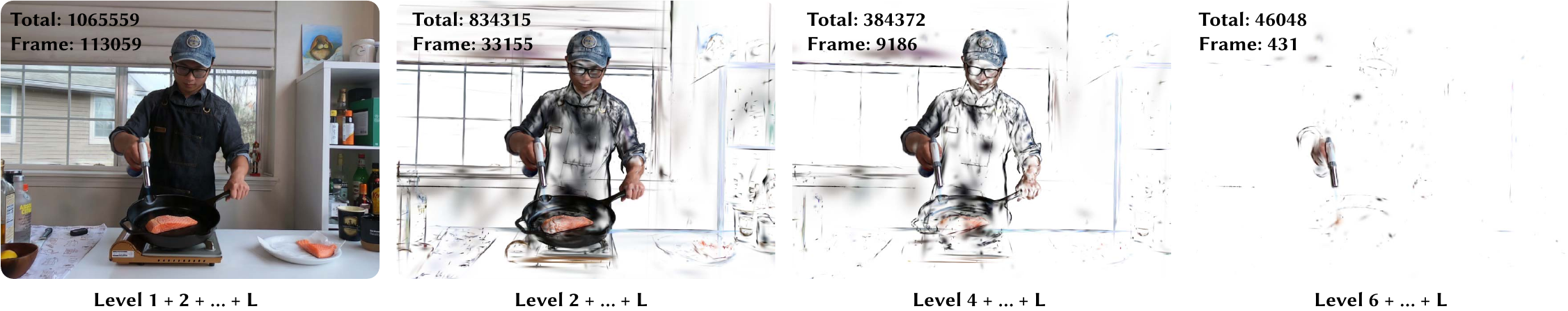}
    \caption{
        \textbf{Visual results for different levels of \tgh on the 1200-frame Neural3DV \cite{li2022neural} dataset}.
        \camera{The number of 4D Gaussians in each level range is shown on the top left corner. }
        \camera{Note that the as the level increases, \revision{the movement speed of the Gaussians increase accordingly.}}
    }
    \label{fig:ablation-level-results}
\end{figure*}

\subsubsection{Ablations on \tgh}
\ \\
We ablate the hyperparameters for \tgh and analyze the effects of \tgh as the frames of video increase.

\paragraph{Number of Levels}
We demonstrate quantitative comparisons of VRAM usage and iteration speed for our models trained with varying numbers of levels in Fig. \ref{fig:ablation-tgh} (a).
It is important to note that our model with only 1 level will degrade into the conventional 4DGS  with our \cpm, resulting in increased VRAM usage and slower iteration speed during training.
We observe that our model consistently achieves better efficiency as the number of levels increases.
However, using a too-large level count introduces overhead in the sampling and update process of \tgh, resulting in lower iteration speed.
Therefore, we use 9 levels as the default setting to train our model.
Additionally, we include visual results for different levels of TGH.
As shown in \cref{fig:ablation-level-results}, the coarse level  represents static scenes like the background, while the fine level with short temporal segments is used to model objects with dynamic motions.

\paragraph{Root Segment Length}
The root segment length is also a key factor affecting the computational efficiency of \tgh.
We conduct a comprehensive study with different lengths of the root segment.
Fig. \ref{fig:ablation-tgh} (b) showcases that by increasing the segment length, the training cost initially decreases and then increases significantly.
This is because when the segment length is either too long or too short, Gaussians influencing vastly different timestamps may be assigned to the same segment or placed into the global segment (\cref{sec:implementation_details}), either due to the shortest segment being too long or the longest segment being too short.
This increases the probability of sampling Gaussians that do not contribute to rendering at specific timestamps, resulting in an increase in both VRAM usage and iteration time.
From our experiments, we found that a length of 10$s$ is optimal, and we chose it as the default setting for our experiments.

\paragraph{Video Frames}
As discussed in \cref{sec:temporal_gaussian_hierarchy}, one of the strengths of \tgh is its nearly constant training cost regardless of the duration of video data.
We evaluate the training cost of models with and without \tgh as the number of training frames increases.
As illustrated in Fig. \ref{fig:ablation-tgh} (c), our \tgh structure maintains a constant VRAM cost and iteration speed even with very long video frames, whereas the model without \tgh incurs significantly higher computational costs.
This scalability facilitates the training of long volumetric videos using our methods.

\begin{figure}
    \includegraphics[width=0.9\linewidth]{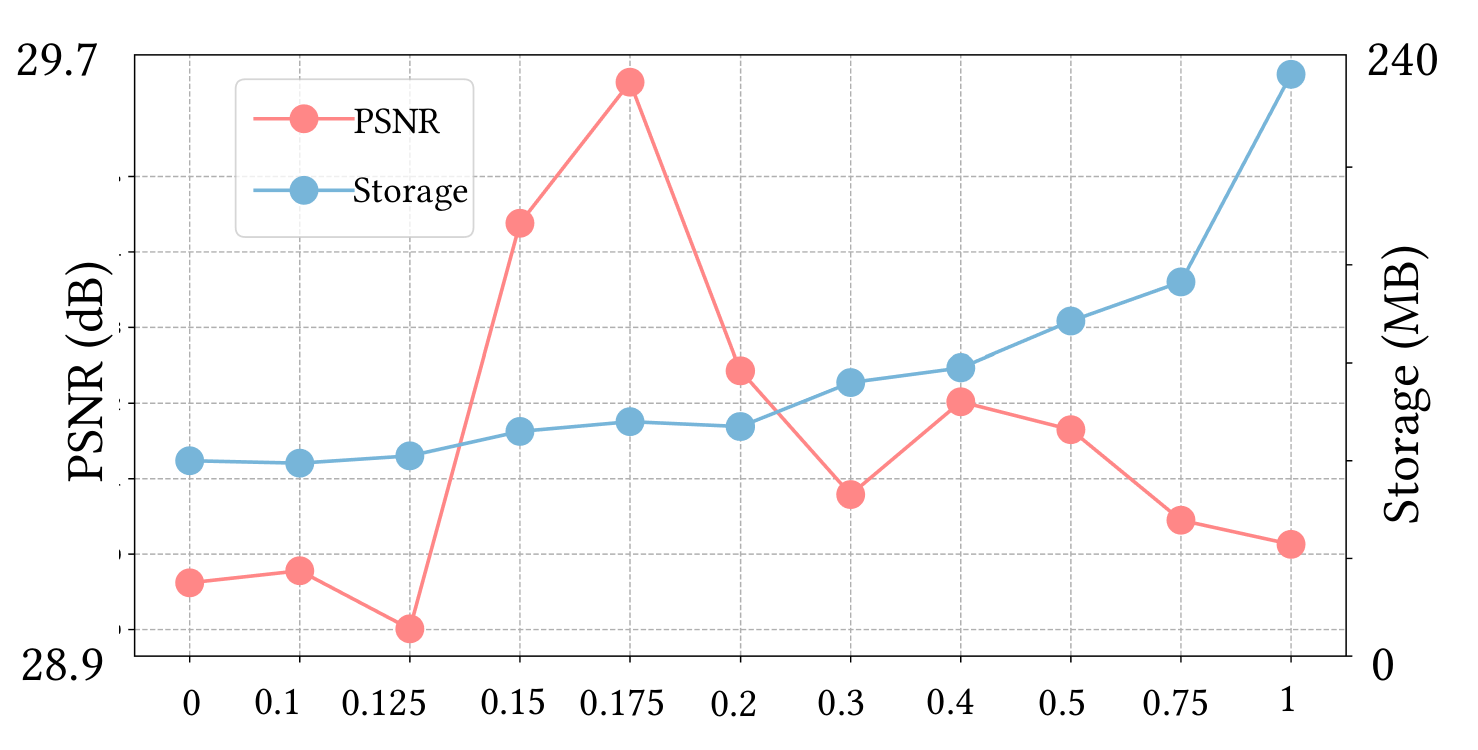}
    \caption{
        \textbf{Comparison on Storage cost and Image Quality with different ratio parameters.} \camera{\revision{Storage cost decreases as the ratio drops.} However, increasing the ratio does not necessarily lead to better results, as the model may overfit the scene. Based on these observations, we choose $\lambda_h$=0.15 for our method to balance storage efficiency and image quality.}}
    \label{fig:ablation-gradthr}
\end{figure}

\begin{figure*}
    \includegraphics[width=1.0\linewidth]{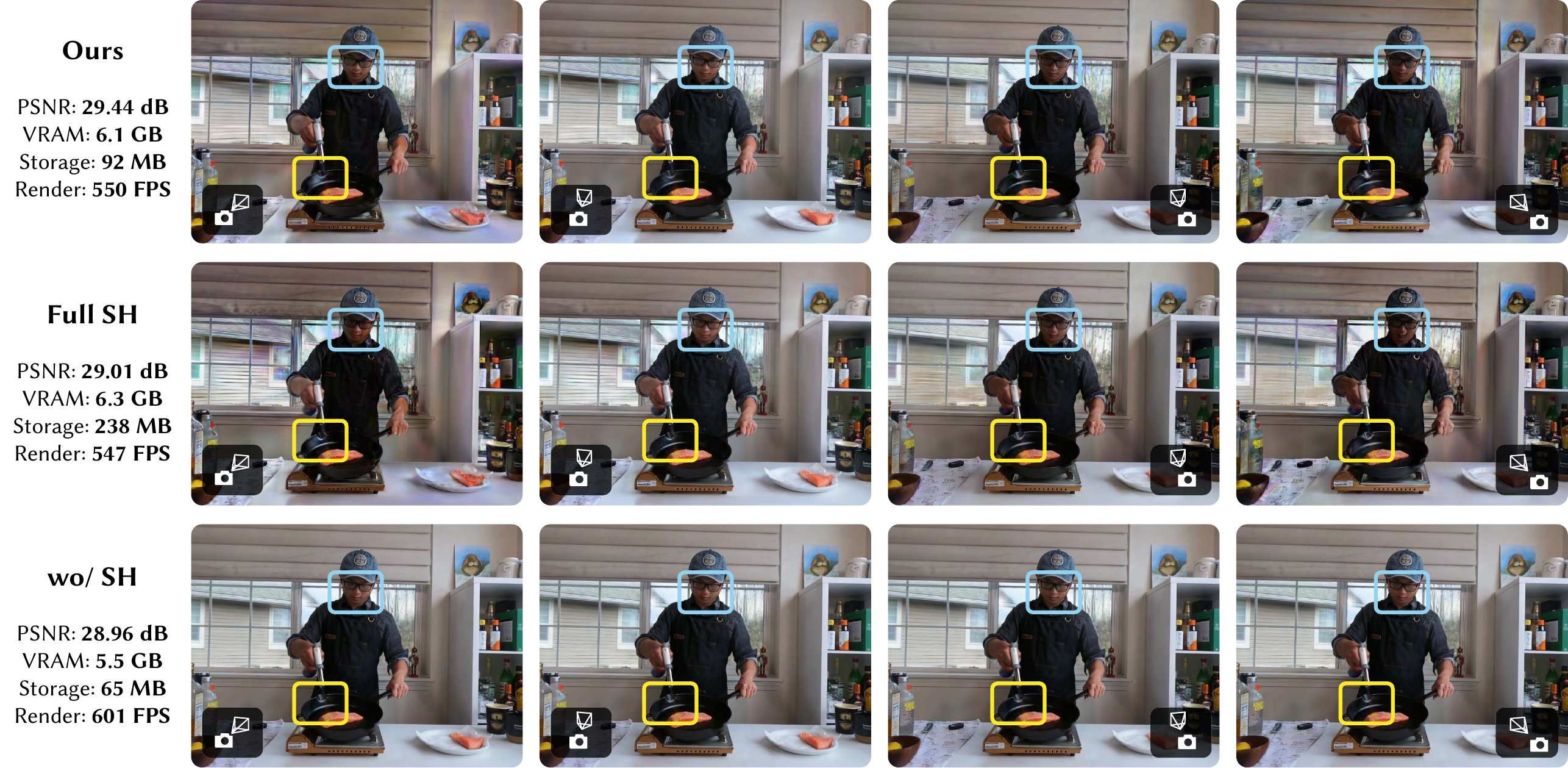}
    \caption{\textbf{Qualitative comparisons for ablations on \cpm.} We compare the view-dependent effects between our method and the model with full SH as well as without SH. \camera{Our proposed Compact Appearance Model successfully captures the view-dependency of the pan and face region.} }
    \label{fig:ablation-viewdependent}
\end{figure*}

\subsubsection{Ablations on Appearance Model}
\begin{table}[t!]
  \caption{\textbf{Comparison on Storage and Image Quality with different gradient thresholds.} Low threshold leads to all residual SH $\mathbf{h}$ being enabled early, which might lead to slightly worse rendering quality, while large values might identify too small amount of view-dependent points.}
  \label{tab:ablation-gradient}
  \resizebox{0.85\linewidth}{!}{%
    \begin{tabular}{l|ccccc}
      \toprule
                           & 0              & $1e^{-7}$ & $1e^{-6}$ (Ours) & $1e^{-3}$   & $\infty$      \\
      \midrule
      Storage (MB)         & 238            & 94        & 92               & \textbf{65} & \textbf{65} \\
      PSNR  ${\uparrow}$   & 29.01          & 28.93     & \textbf{29.44}   & 29.06       & 28.96       \\
      SSIM  ${\uparrow}$   & 0.953          & 0.953     & \textbf{0.954}   & 0.951       & 0.950       \\
      LPIPS ${\downarrow}$ & \textbf{0.214} & 0.215     & \textbf{0.214}   & 0.217       & 0.219       \\
      \bottomrule
    \end{tabular}
  }
\end{table}

We analyze the impact of the ratio parameter, $\lambda_{\mathrm{h}}$, in the appearance model.
We assess the disk storage of models with varying gradient ratios.
Notably, when the ratio is 0, only the base color is utilized to model diffuse appearance, while a ratio of 1 indicates that we enable all SH coefficients to model view-dependent effects.
As illustrated in Fig. \ref{fig:ablation-gradthr}, we observe that the storage cost decreases as the ratio decreases.
However, this reduction in storage comes at the expense of weakening the ability to model view-dependent effects, resulting in a decrease in image quality.
Experimental results indicate that using a ratio of 0.15 can significantly reduce storage consumption without compromising rendering results.
We also analyze the effect of different gradient thresholds when enabling residual SH, as shown in Tab. \cref{tab:ablation-gradient}. According to the experiments, we select a gradient threshold of 0.000001 for our method.

We also compare the results of view-dependent effects for different models in Fig. \ref{fig:ablation-viewdependent}.
When the SH coefficients are zero, the rendering results only exhibit a diffuse appearance without any view-dependent effects.
Our sparse SH representation retains the ability to express view-dependent effects similar to full SH while significantly reducing storage space, demonstrating the effectiveness of our \cpm.

\camera{
    \subsubsection{Ablations on the Hardware Rasterizer}
    \begin{table}[t!]
    \caption{
        \camera{
            \textbf{Comparison on the sorting speed using different backends on an RTX 4090 and i9-13900K.}
            Sorting directly on the GPU using CUDA \cite{paszke2019pytorch} proves to be much more performant compared to the CPU-sorting \cite{mkkellogg2024GaussianSplats3D} or \revision{Compute-Shader-sorting \cite{3dgstutorial} implementation for the graphics-pipeline-based rasterizer.}
        }
    }
    \label{tab:ablation-sorting}
    \resizebox{0.9\linewidth}{!}{%
        \begin{tabular}{l|ccc}
            \toprule
            \revision{Sorting Backend}               & \camera{ CUDA (Ours)}          & \camera{CPU   } & \camera{Compute Shader} \\
            \midrule
            \camera{Million-Points / s ${\uparrow}$} & \camera{ \textbf{8844}       } & \camera{16.28 } & \camera{152.2}          \\
            \bottomrule
        \end{tabular}
    }
\end{table}

    In addtion to comparing the rendering speed of the proposed hardware rasterizer \cite{shreiner2009opengl} to the original software rasterizer \cite{kerbl20233d} in \cref{tab:ablation-components}, we also
    evaluate the impact of different sorting backends for the hardware rasterizer in \cref{tab:ablation-sorting}.
    This ablation is done on a high-end consumer machine with an RTX 4090 GPU and an i9-13900K CPU.
    As shown in the table, using the CUDA backend \cite{paszke2019pytorch} greatly increases GPU utilization and sorting throughput and aids in our increased rendering speed compared to the CPU-sorting \cite{mkkellogg2024GaussianSplats3D} or \revision{Compute-Shader-sorting backend \cite{3dgstutorial} for the graphics-pipeline-based rasterizer}.
}

\begin{figure}
    \includegraphics[width=1.0\linewidth]{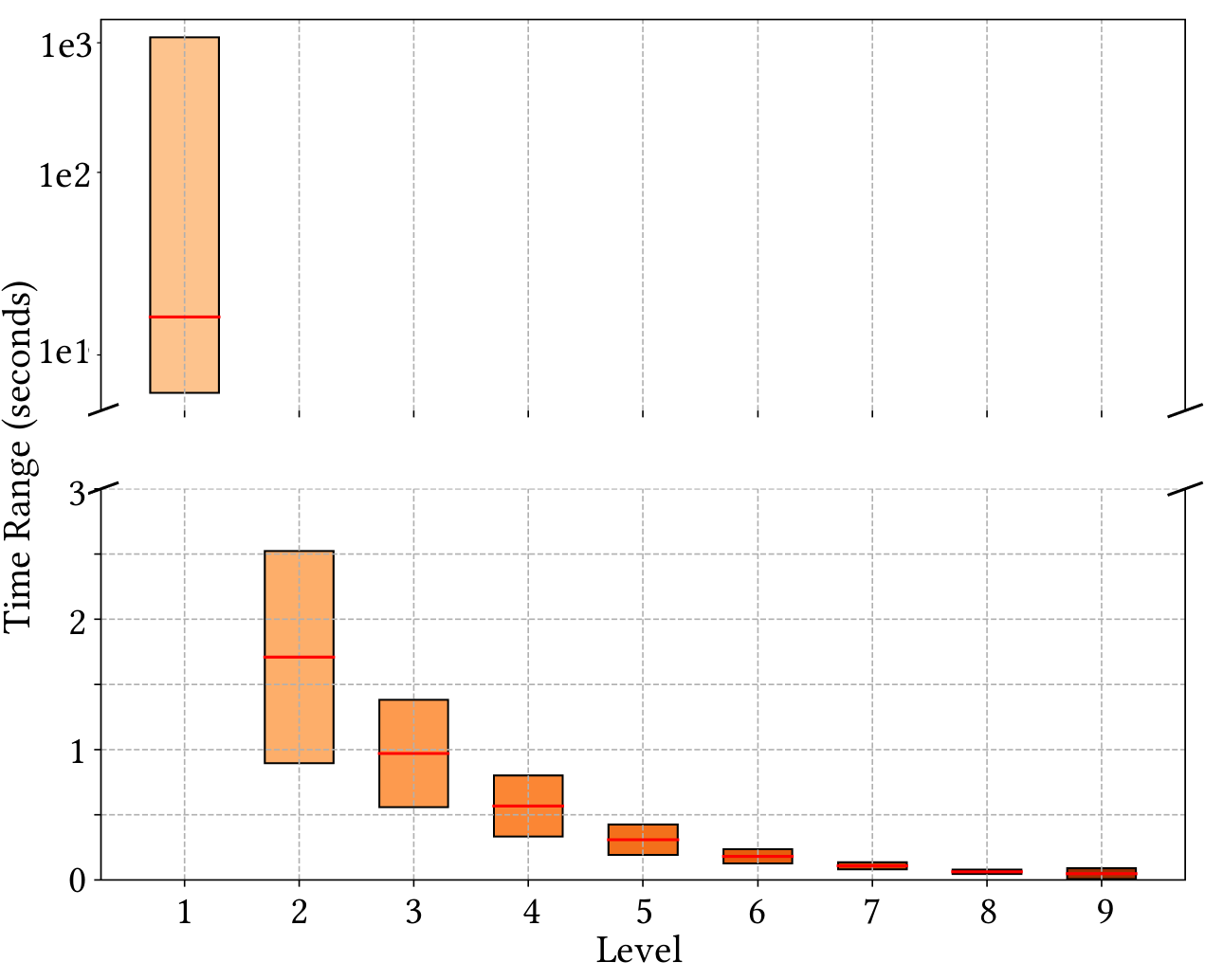}
    \caption{
        \camera{
            \textbf{Visualization on the mean and standard deviation of the range of the 4D Gaussians on the 1200-frame (40s) \textit{flame\_salmon} sequence of the \textit{Neural3DV} dataset determined by \cref{eq:gaussian_influence_range}.}
        }
    }
    \label{fig:ablation-meanstd}
\end{figure}

\begin{figure}
    \includegraphics[width=0.985\linewidth]{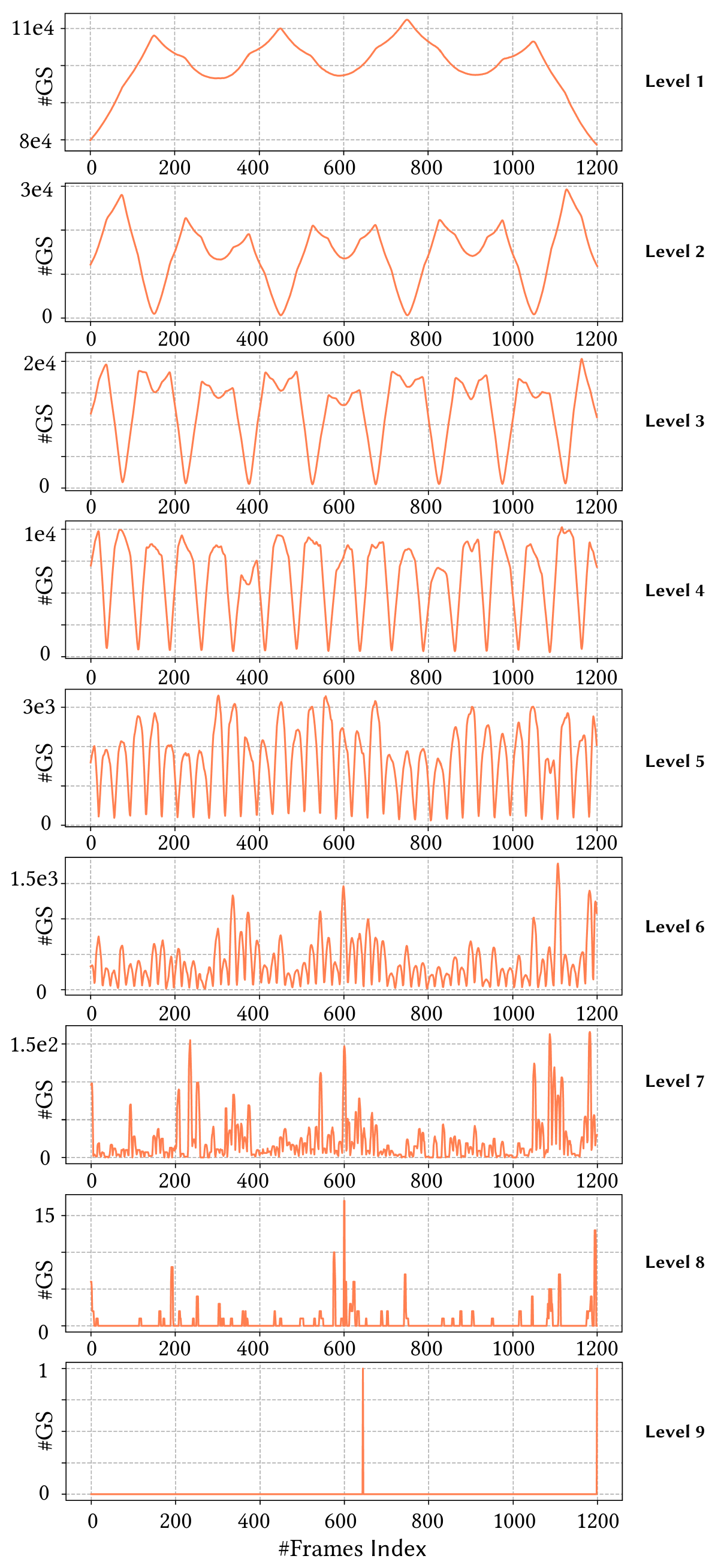}
    \caption{
        \camera{
            \textbf{Visualization on the number of 4D Gaussians for each of the 9 levels across the 1200-frame (40s) duration of the \textit{flame\_salmon} sequence of the \textit{Neural3DV} dataset.}
        }
    }
    \label{fig:ablation-times}
\end{figure}

\camera{

    \subsubsection{Visualization on the Distribution of 4D Gaussians}
    In \cref{fig:ablation-meanstd}, \cref{fig:ablation-times} and \cref{fig:ablation-level-results} we visualize the distribution of 4D Gaussians in each level of the \tgh across time and addtionally show the rendering results of different levels.
    As shown by the figures, although the division of segments is uniform in the \tgh, the number of 4D Gaussians in each segment is different, achieving a non-uniform subdivision of representational power across time.
}

\section{Conclusion}

We proposed temporal Gaussian hierarchy, a novel 4D scene representation to reconstruct long volumetric videos with low storage cost from multi-view RGB videos.
The representation is built based on the observation that dynamic scenes typically exhibit varying degrees of temporal redundancy, as there are generally scene areas with different motions.
Specifically, the Gaussian hierarchy consists of multiple levels, and each level independently describes regions of the scene with different degrees of content change based on a set of segments.
Our approach stores a set of Gaussian primitives in these segments and adaptively shares them to depict the scene content across various temporal scales.
We further exploited the tree-like structure of the hierarchy to efficiently represent the scene at a time step using a subset of Gaussian primitives, resulting in nearly constant GPU memory usage during the training or rendering process, irrespective of the volumetric video's length.
In addition, we designed a \cpm that integrated diffuse and view-dependent Gaussians to reduce the model size.
We also developed a rasterization pipeline for Gaussian primitives leveraging hardware-accelerated techniques to enhance rendering efficiency.
In conclusion, both qualitative and quantitative evaluations confirm that our method surpasses existing baselines, offering state-of-the-art visual quality and rendering speed. Moreover, our method requires significantly less training cost and memory usage, facilitating the reconstruction and rendering of long volumetric videos.

\section{Discussion}

\paragraph{Limitation} Although our work shows significant improvement on long volumetric videos, it still has some limitations.
First, our reconstruction system is not real-time, requiring several hours to transform volumetric videos into our 4D representation. To further improve training efficiency, we could instill a stronger geometry prior or regularize the distribution of Gaussian primitives.
Moreover, our method requires semi-dense views to adequately cover dynamic scenes, and it doesn't generalize well in a sparse-view setting.
Inspired by prior works~\cite{wu2023reconfusion}, we plan to use the strong generative priors in diffusion models~\cite{ldm} to improve sparse-view reconstruction results.

\begin{acks}
    The authors would like to acknowledge support from NSFC (No. 62172364), Information Technology Center, and State Key Lab of CAD\&CG, Zhejiang University.
\end{acks}

\bibliographystyle{ACM-Reference-Format}
\bibliography{reference}


\begin{thebibliography}{77}


\ifx \showCODEN    \undefined \def \showCODEN     #1{\unskip}     \fi
\ifx \showDOI      \undefined \def \showDOI       #1{#1}\fi
\ifx \showISBNx    \undefined \def \showISBNx     #1{\unskip}     \fi
\ifx \showISBNxiii \undefined \def \showISBNxiii  #1{\unskip}     \fi
\ifx \showISSN     \undefined \def \showISSN      #1{\unskip}     \fi
\ifx \showLCCN     \undefined \def \showLCCN      #1{\unskip}     \fi
\ifx \shownote     \undefined \def \shownote      #1{#1}          \fi
\ifx \showarticletitle \undefined \def \showarticletitle #1{#1}   \fi
\ifx \showURL      \undefined \def \showURL       {\relax}        \fi
\providecommand\bibfield[2]{#2}
\providecommand\bibinfo[2]{#2}
\providecommand\natexlab[1]{#1}
\providecommand\showeprint[2][]{arXiv:#2}

\bibitem[Ahmed et~al\mbox{.}(2008)]%
        {ahmed2008dense}
\bibfield{author}{\bibinfo{person}{Naveed Ahmed}, \bibinfo{person}{Christian Theobalt}, \bibinfo{person}{Christian Rossl}, \bibinfo{person}{Sebastian Thrun}, {and} \bibinfo{person}{Hans-Peter Seidel}.} \bibinfo{year}{2008}\natexlab{}.
\newblock \showarticletitle{Dense correspondence finding for parametrization-free animation reconstruction from video}.
\newblock


\bibitem[Aliev et~al\mbox{.}(2020)]%
        {aliev2020neural}
\bibfield{author}{\bibinfo{person}{Kara-Ali Aliev}, \bibinfo{person}{Artem Sevastopolsky}, \bibinfo{person}{Maria Kolos}, \bibinfo{person}{Dmitry Ulyanov}, {and} \bibinfo{person}{Victor Lempitsky}.} \bibinfo{year}{2020}\natexlab{}.
\newblock \showarticletitle{Neural point-based graphics}. In \bibinfo{booktitle}{\emph{ECCV}}.
\newblock


\bibitem[Bozic et~al\mbox{.}(2020)]%
        {bozic2020deepdeform}
\bibfield{author}{\bibinfo{person}{Aljaz Bozic}, \bibinfo{person}{Michael Zollhofer}, \bibinfo{person}{Christian Theobalt}, {and} \bibinfo{person}{Matthias Nie{\ss}ner}.} \bibinfo{year}{2020}\natexlab{}.
\newblock \showarticletitle{Deepdeform: Learning non-rigid rgb-d reconstruction with semi-supervised data}. In \bibinfo{booktitle}{\emph{CVPR}}.
\newblock


\bibitem[Cao and Johnson(2023)]%
        {Cao2023HEXPLANE}
\bibfield{author}{\bibinfo{person}{Ang Cao} {and} \bibinfo{person}{Justin Johnson}.} \bibinfo{year}{2023}\natexlab{}.
\newblock \showarticletitle{HexPlane: A Fast Representation for Dynamic Scenes}.
\newblock  (\bibinfo{year}{2023}).
\newblock


\bibitem[Carranza et~al\mbox{.}(2003)]%
        {carranza2003free}
\bibfield{author}{\bibinfo{person}{Joel Carranza}, \bibinfo{person}{Christian Theobalt}, \bibinfo{person}{Marcus~A Magnor}, {and} \bibinfo{person}{Hans-Peter Seidel}.} \bibinfo{year}{2003}\natexlab{}.
\newblock \showarticletitle{Free-viewpoint video of human actors}.
\newblock \bibinfo{journal}{\emph{ACM transactions on graphics (TOG)}} \bibinfo{volume}{22}, \bibinfo{number}{3} (\bibinfo{year}{2003}), \bibinfo{pages}{569--577}.
\newblock


\bibitem[Chen et~al\mbox{.}(2022)]%
        {chen2022tensorf}
\bibfield{author}{\bibinfo{person}{Anpei Chen}, \bibinfo{person}{Zexiang Xu}, \bibinfo{person}{Andreas Geiger}, \bibinfo{person}{Jingyi Yu}, {and} \bibinfo{person}{Hao Su}.} \bibinfo{year}{2022}\natexlab{}.
\newblock \showarticletitle{Tensorf: Tensorial radiance fields}. In \bibinfo{booktitle}{\emph{European Conference on Computer Vision}}. Springer.
\newblock


\bibitem[Chen et~al\mbox{.}(2021)]%
        {chen2021mvsnerf}
\bibfield{author}{\bibinfo{person}{Anpei Chen}, \bibinfo{person}{Zexiang Xu}, \bibinfo{person}{Fuqiang Zhao}, \bibinfo{person}{Xiaoshuai Zhang}, \bibinfo{person}{Fanbo Xiang}, \bibinfo{person}{Jingyi Yu}, {and} \bibinfo{person}{Hao Su}.} \bibinfo{year}{2021}\natexlab{}.
\newblock \showarticletitle{Mvsnerf: Fast generalizable radiance field reconstruction from multi-view stereo}. In \bibinfo{booktitle}{\emph{ICCV}}.
\newblock


\bibitem[Chen et~al\mbox{.}(2023)]%
        {chen2023mobilenerf}
\bibfield{author}{\bibinfo{person}{Zhiqin Chen}, \bibinfo{person}{Thomas Funkhouser}, \bibinfo{person}{Peter Hedman}, {and} \bibinfo{person}{Andrea Tagliasacchi}.} \bibinfo{year}{2023}\natexlab{}.
\newblock \showarticletitle{Mobilenerf: Exploiting the polygon rasterization pipeline for efficient neural field rendering on mobile architectures}. In \bibinfo{booktitle}{\emph{CVPR}}.
\newblock


\bibitem[Collet et~al\mbox{.}(2015)]%
        {collet2015hqfvv}
\bibfield{author}{\bibinfo{person}{Alvaro Collet}, \bibinfo{person}{Ming Chuang}, \bibinfo{person}{Pat Sweeney}, \bibinfo{person}{Don Gillett}, \bibinfo{person}{Dennis Evseev}, \bibinfo{person}{David Calabrese}, \bibinfo{person}{Hugues Hoppe}, \bibinfo{person}{Adam Kirk}, {and} \bibinfo{person}{Steve Sullivan}.} \bibinfo{year}{2015}\natexlab{}.
\newblock \showarticletitle{High-quality streamable free-viewpoint video}.
\newblock \bibinfo{journal}{\emph{ACM Transactions on Graphics (ToG)}} \bibinfo{volume}{34}, \bibinfo{number}{4} (\bibinfo{year}{2015}), \bibinfo{pages}{1--13}.
\newblock


\bibitem[Du et~al\mbox{.}(2021)]%
        {du2021neural}
\bibfield{author}{\bibinfo{person}{Yilun Du}, \bibinfo{person}{Yinan Zhang}, \bibinfo{person}{Hong-Xing Yu}, \bibinfo{person}{Joshua~B Tenenbaum}, {and} \bibinfo{person}{Jiajun Wu}.} \bibinfo{year}{2021}\natexlab{}.
\newblock \showarticletitle{Neural radiance flow for 4d view synthesis and video processing. In 2021 IEEE}. In \bibinfo{booktitle}{\emph{CVF International Conference on Computer Vision (ICCV)}}. \bibinfo{pages}{14304--14314}.
\newblock


\bibitem[Duan et~al\mbox{.}(2024)]%
        {duan20244d}
\bibfield{author}{\bibinfo{person}{Yuanxing Duan}, \bibinfo{person}{Fangyin Wei}, \bibinfo{person}{Qiyu Dai}, \bibinfo{person}{Yuhang He}, \bibinfo{person}{Wenzheng Chen}, {and} \bibinfo{person}{Baoquan Chen}.} \bibinfo{year}{2024}\natexlab{}.
\newblock \showarticletitle{4D Gaussian Splatting: Towards Efficient Novel View Synthesis for Dynamic Scenes}.
\newblock \bibinfo{journal}{\emph{arXiv preprint arXiv:2402.03307}} (\bibinfo{year}{2024}).
\newblock


\bibitem[Fang et~al\mbox{.}(2022)]%
        {fang2022fast}
\bibfield{author}{\bibinfo{person}{Jiemin Fang}, \bibinfo{person}{Taoran Yi}, \bibinfo{person}{Xinggang Wang}, \bibinfo{person}{Lingxi Xie}, \bibinfo{person}{Xiaopeng Zhang}, \bibinfo{person}{Wenyu Liu}, \bibinfo{person}{Matthias Nie{\ss}ner}, {and} \bibinfo{person}{Qi Tian}.} \bibinfo{year}{2022}\natexlab{}.
\newblock \showarticletitle{Fast dynamic radiance fields with time-aware neural voxels}. In \bibinfo{booktitle}{\emph{SIGGRAPH Asia 2022 Conference Papers}}.
\newblock


\bibitem[Fridovich-Keil et~al\mbox{.}(2023)]%
        {kplanes_2023}
\bibfield{author}{\bibinfo{person}{Sara Fridovich-Keil}, \bibinfo{person}{Giacomo Meanti}, \bibinfo{person}{Frederik~Rahbæk Warburg}, \bibinfo{person}{Benjamin Recht}, {and} \bibinfo{person}{Angjoo Kanazawa}.} \bibinfo{year}{2023}\natexlab{}.
\newblock \showarticletitle{K-Planes: Explicit Radiance Fields in Space, Time, and Appearance}. In \bibinfo{booktitle}{\emph{CVPR}}.
\newblock


\bibitem[Fridovich-Keil et~al\mbox{.}(2022)]%
        {fridovich2022plenoxels}
\bibfield{author}{\bibinfo{person}{Sara Fridovich-Keil}, \bibinfo{person}{Alex Yu}, \bibinfo{person}{Matthew Tancik}, \bibinfo{person}{Qinhong Chen}, \bibinfo{person}{Benjamin Recht}, {and} \bibinfo{person}{Angjoo Kanazawa}.} \bibinfo{year}{2022}\natexlab{}.
\newblock \showarticletitle{Plenoxels: Radiance fields without neural networks}. In \bibinfo{booktitle}{\emph{CVPR}}.
\newblock


\bibitem[Garbin et~al\mbox{.}(2021)]%
        {garbin2021fastnerf}
\bibfield{author}{\bibinfo{person}{Stephan~J Garbin}, \bibinfo{person}{Marek Kowalski}, \bibinfo{person}{Matthew Johnson}, \bibinfo{person}{Jamie Shotton}, {and} \bibinfo{person}{Julien Valentin}.} \bibinfo{year}{2021}\natexlab{}.
\newblock \showarticletitle{Fastnerf: High-fidelity neural rendering at 200fps}.
\newblock


\bibitem[Hasselgren et~al\mbox{.}(2022)]%
        {hasselgren2022shape}
\bibfield{author}{\bibinfo{person}{Jon Hasselgren}, \bibinfo{person}{Nikolai Hofmann}, {and} \bibinfo{person}{Jacob Munkberg}.} \bibinfo{year}{2022}\natexlab{}.
\newblock \showarticletitle{Shape, light, and material decomposition from images using monte carlo rendering and denoising}.
\newblock \bibinfo{journal}{\emph{NeuRIPS}} (\bibinfo{year}{2022}).
\newblock


\bibitem[Hedman et~al\mbox{.}(2021)]%
        {hedman2021baking}
\bibfield{author}{\bibinfo{person}{Peter Hedman}, \bibinfo{person}{Pratul~P Srinivasan}, \bibinfo{person}{Ben Mildenhall}, \bibinfo{person}{Jonathan~T Barron}, {and} \bibinfo{person}{Paul Debevec}.} \bibinfo{year}{2021}\natexlab{}.
\newblock \showarticletitle{Baking neural radiance fields for real-time view synthesis}.
\newblock


\bibitem[Huffman(1952)]%
        {huffman1952method}
\bibfield{author}{\bibinfo{person}{David~A Huffman}.} \bibinfo{year}{1952}\natexlab{}.
\newblock \showarticletitle{A method for the construction of minimum-redundancy codes}.
\newblock \bibinfo{journal}{\emph{Proceedings of the IRE}} \bibinfo{volume}{40}, \bibinfo{number}{9} (\bibinfo{year}{1952}), \bibinfo{pages}{1098--1101}.
\newblock


\bibitem[Joo et~al\mbox{.}(2015)]%
        {joo2015panoptic}
\bibfield{author}{\bibinfo{person}{Hanbyul Joo}, \bibinfo{person}{Hao Liu}, \bibinfo{person}{Lei Tan}, \bibinfo{person}{Lin Gui}, \bibinfo{person}{Bart Nabbe}, \bibinfo{person}{Iain Matthews}, \bibinfo{person}{Takeo Kanade}, \bibinfo{person}{Shohei Nobuhara}, {and} \bibinfo{person}{Yaser Sheikh}.} \bibinfo{year}{2015}\natexlab{}.
\newblock \showarticletitle{Panoptic studio: A massively multiview system for social motion capture}. In \bibinfo{booktitle}{\emph{Proceedings of the IEEE international conference on computer vision}}. \bibinfo{pages}{3334--3342}.
\newblock


\bibitem[Jung et~al\mbox{.}(2023)]%
        {jung2023deformable}
\bibfield{author}{\bibinfo{person}{HyunJun Jung}, \bibinfo{person}{Nikolas Brasch}, \bibinfo{person}{Jifei Song}, \bibinfo{person}{Eduardo Perez-Pellitero}, \bibinfo{person}{Yiren Zhou}, \bibinfo{person}{Zhihao Li}, \bibinfo{person}{Nassir Navab}, {and} \bibinfo{person}{Benjamin Busam}.} \bibinfo{year}{2023}\natexlab{}.
\newblock \showarticletitle{Deformable 3d gaussian splatting for animatable human avatars}.
\newblock \bibinfo{journal}{\emph{arXiv preprint arXiv:2312.15059}} (\bibinfo{year}{2023}).
\newblock


\bibitem[Kerbl et~al\mbox{.}(2023)]%
        {kerbl20233d}
\bibfield{author}{\bibinfo{person}{Bernhard Kerbl}, \bibinfo{person}{Georgios Kopanas}, \bibinfo{person}{Thomas Leimk{\"u}hler}, {and} \bibinfo{person}{George Drettakis}.} \bibinfo{year}{2023}\natexlab{}.
\newblock \showarticletitle{3D gaussian splatting for real-time radiance field rendering}.
\newblock \bibinfo{journal}{\emph{ACM Transactions on Graphics (TOG)}} \bibinfo{volume}{42}, \bibinfo{number}{4} (\bibinfo{year}{2023}), \bibinfo{pages}{1--14}.
\newblock


\bibitem[Kingma and Ba(2014)]%
        {kingma2014adam}
\bibfield{author}{\bibinfo{person}{Diederik~P Kingma} {and} \bibinfo{person}{Jimmy Ba}.} \bibinfo{year}{2014}\natexlab{}.
\newblock \showarticletitle{Adam: A method for stochastic optimization}.
\newblock \bibinfo{journal}{\emph{arXiv preprint arXiv:1412.6980}} (\bibinfo{year}{2014}).
\newblock


\bibitem[Kopanas et~al\mbox{.}(2024)]%
        {3dgstutorial}
\bibfield{author}{\bibinfo{person}{Georgios Kopanas}, \bibinfo{person}{Bernhard Kerbl}, \bibinfo{person}{Antoine Guédon}, {and} \bibinfo{person}{Jonathon Luiten}.} \bibinfo{year}{2024}\natexlab{}.
\newblock \bibinfo{title}{{3D Gaussian Splatting Tutorial}}.
\newblock
\newblock
\urldef\tempurl%
\url{https://3dgstutorial.github.io/}
\showURL{%
\tempurl}
\newblock
\shownote{International Conference on 3D Vision Tutorial}.


\bibitem[Krizhevsky et~al\mbox{.}(2012)]%
        {krizhevsky2012imagenet}
\bibfield{author}{\bibinfo{person}{Alex Krizhevsky}, \bibinfo{person}{Ilya Sutskever}, {and} \bibinfo{person}{Geoffrey~E Hinton}.} \bibinfo{year}{2012}\natexlab{}.
\newblock \showarticletitle{Imagenet classification with deep convolutional neural networks}.
\newblock \bibinfo{journal}{\emph{Advances in neural information processing systems}}  \bibinfo{volume}{25} (\bibinfo{year}{2012}).
\newblock


\bibitem[Kulhanek and Sattler(2023)]%
        {kulhanek2023tetra}
\bibfield{author}{\bibinfo{person}{Jonas Kulhanek} {and} \bibinfo{person}{Torsten Sattler}.} \bibinfo{year}{2023}\natexlab{}.
\newblock \showarticletitle{Tetra-nerf: Representing neural radiance fields using tetrahedra}. In \bibinfo{booktitle}{\emph{ICCV}}.
\newblock


\bibitem[Lassner and Zollhofer(2021)]%
        {lassner2021pulsar}
\bibfield{author}{\bibinfo{person}{Christoph Lassner} {and} \bibinfo{person}{Michael Zollhofer}.} \bibinfo{year}{2021}\natexlab{}.
\newblock \showarticletitle{Pulsar: Efficient sphere-based neural rendering}. In \bibinfo{booktitle}{\emph{CVPR}}.
\newblock


\bibitem[Li et~al\mbox{.}(2022)]%
        {li2022neural}
\bibfield{author}{\bibinfo{person}{Tianye Li}, \bibinfo{person}{Mira Slavcheva}, \bibinfo{person}{Michael Zollhoefer}, \bibinfo{person}{Simon Green}, \bibinfo{person}{Christoph Lassner}, \bibinfo{person}{Changil Kim}, \bibinfo{person}{Tanner Schmidt}, \bibinfo{person}{Steven Lovegrove}, \bibinfo{person}{Michael Goesele}, \bibinfo{person}{Richard Newcombe}, {et~al\mbox{.}}} \bibinfo{year}{2022}\natexlab{}.
\newblock \showarticletitle{Neural 3d video synthesis from multi-view video}. In \bibinfo{booktitle}{\emph{Proceedings of the IEEE/CVF Conference on Computer Vision and Pattern Recognition}}. \bibinfo{pages}{5521--5531}.
\newblock


\bibitem[Li et~al\mbox{.}(2023)]%
        {li2023dynibar}
\bibfield{author}{\bibinfo{person}{Zhengqi Li}, \bibinfo{person}{Qianqian Wang}, \bibinfo{person}{Forrester Cole}, \bibinfo{person}{Richard Tucker}, {and} \bibinfo{person}{Noah Snavely}.} \bibinfo{year}{2023}\natexlab{}.
\newblock \showarticletitle{Dynibar: Neural dynamic image-based rendering}. In \bibinfo{booktitle}{\emph{Proceedings of the IEEE/CVF Conference on Computer Vision and Pattern Recognition}}. \bibinfo{pages}{4273--4284}.
\newblock


\bibitem[Lin et~al\mbox{.}(2023b)]%
        {lin2023im4d}
\bibfield{author}{\bibinfo{person}{Haotong Lin}, \bibinfo{person}{Sida Peng}, \bibinfo{person}{Zhen Xu}, \bibinfo{person}{Tao Xie}, \bibinfo{person}{Xingyi He}, \bibinfo{person}{Hujun Bao}, {and} \bibinfo{person}{Xiaowei Zhou}.} \bibinfo{year}{2023}\natexlab{b}.
\newblock \showarticletitle{High-Fidelity and Real-Time Novel View Synthesis for Dynamic Scenes}. In \bibinfo{booktitle}{\emph{SIGGRAPH Asia Conference Proceedings}}.
\newblock


\bibitem[Lin et~al\mbox{.}(2022)]%
        {lin2022efficient}
\bibfield{author}{\bibinfo{person}{Haotong Lin}, \bibinfo{person}{Sida Peng}, \bibinfo{person}{Zhen Xu}, \bibinfo{person}{Yunzhi Yan}, \bibinfo{person}{Qing Shuai}, \bibinfo{person}{Hujun Bao}, {and} \bibinfo{person}{Xiaowei Zhou}.} \bibinfo{year}{2022}\natexlab{}.
\newblock \showarticletitle{Efficient Neural Radiance Fields for Interactive Free-viewpoint Video}. In \bibinfo{booktitle}{\emph{SIGGRAPH Asia Conference Proceedings}}.
\newblock


\bibitem[Lin et~al\mbox{.}(2023a)]%
        {lin2023gaussian}
\bibfield{author}{\bibinfo{person}{Youtian Lin}, \bibinfo{person}{Zuozhuo Dai}, \bibinfo{person}{Siyu Zhu}, {and} \bibinfo{person}{Yao Yao}.} \bibinfo{year}{2023}\natexlab{a}.
\newblock \showarticletitle{Gaussian-flow: 4d reconstruction with dynamic 3d gaussian particle}.
\newblock \bibinfo{journal}{\emph{arXiv preprint arXiv:2312.03431}} (\bibinfo{year}{2023}).
\newblock


\bibitem[Ling et~al\mbox{.}(2023)]%
        {ling2023align}
\bibfield{author}{\bibinfo{person}{Huan Ling}, \bibinfo{person}{Seung~Wook Kim}, \bibinfo{person}{Antonio Torralba}, \bibinfo{person}{Sanja Fidler}, {and} \bibinfo{person}{Karsten Kreis}.} \bibinfo{year}{2023}\natexlab{}.
\newblock \showarticletitle{Align your gaussians: Text-to-4d with dynamic 3d gaussians and composed diffusion models}.
\newblock \bibinfo{journal}{\emph{arXiv preprint arXiv:2312.13763}} (\bibinfo{year}{2023}).
\newblock


\bibitem[Lombardi et~al\mbox{.}(2019)]%
        {lombardi2019neural}
\bibfield{author}{\bibinfo{person}{Stephen Lombardi}, \bibinfo{person}{Tomas Simon}, \bibinfo{person}{Jason Saragih}, \bibinfo{person}{Gabriel Schwartz}, \bibinfo{person}{Andreas Lehrmann}, {and} \bibinfo{person}{Yaser Sheikh}.} \bibinfo{year}{2019}\natexlab{}.
\newblock \showarticletitle{Neural volumes: Learning dynamic renderable volumes from images}.
\newblock \bibinfo{journal}{\emph{arXiv preprint arXiv:1906.07751}} (\bibinfo{year}{2019}).
\newblock


\bibitem[Long et~al\mbox{.}(2022)]%
        {long2022sparseneus}
\bibfield{author}{\bibinfo{person}{Xiaoxiao Long}, \bibinfo{person}{Cheng Lin}, \bibinfo{person}{Peng Wang}, \bibinfo{person}{Taku Komura}, {and} \bibinfo{person}{Wenping Wang}.} \bibinfo{year}{2022}\natexlab{}.
\newblock \showarticletitle{Sparseneus: Fast generalizable neural surface reconstruction from sparse views}.
\newblock


\bibitem[Lu et~al\mbox{.}(2023)]%
        {lu2023urban}
\bibfield{author}{\bibinfo{person}{Fan Lu}, \bibinfo{person}{Yan Xu}, \bibinfo{person}{Guang Chen}, \bibinfo{person}{Hongsheng Li}, \bibinfo{person}{Kwan-Yee Lin}, {and} \bibinfo{person}{Changjun Jiang}.} \bibinfo{year}{2023}\natexlab{}.
\newblock \showarticletitle{Urban radiance field representation with deformable neural mesh primitives}. In \bibinfo{booktitle}{\emph{ICCV}}.
\newblock


\bibitem[Lu et~al\mbox{.}(2024)]%
        {lu20243d}
\bibfield{author}{\bibinfo{person}{Zhicheng Lu}, \bibinfo{person}{Xiang Guo}, \bibinfo{person}{Le Hui}, \bibinfo{person}{Tianrui Chen}, \bibinfo{person}{Min Yang}, \bibinfo{person}{Xiao Tang}, \bibinfo{person}{Feng Zhu}, {and} \bibinfo{person}{Yuchao Dai}.} \bibinfo{year}{2024}\natexlab{}.
\newblock \showarticletitle{3d geometry-aware deformable gaussian splatting for dynamic view synthesis}.
\newblock \bibinfo{journal}{\emph{arXiv preprint arXiv:2404.06270}} (\bibinfo{year}{2024}).
\newblock


\bibitem[Luiten et~al\mbox{.}(2024)]%
        {luiten2023dynamic}
\bibfield{author}{\bibinfo{person}{Jonathon Luiten}, \bibinfo{person}{Georgios Kopanas}, \bibinfo{person}{Bastian Leibe}, {and} \bibinfo{person}{Deva Ramanan}.} \bibinfo{year}{2024}\natexlab{}.
\newblock \showarticletitle{Dynamic 3D Gaussians: Tracking by Persistent Dynamic View Synthesis}. In \bibinfo{booktitle}{\emph{3DV}}.
\newblock


\bibitem[Mescheder et~al\mbox{.}(2019)]%
        {occnet}
\bibfield{author}{\bibinfo{person}{Lars Mescheder}, \bibinfo{person}{Michael Oechsle}, \bibinfo{person}{Michael Niemeyer}, \bibinfo{person}{Sebastian Nowozin}, {and} \bibinfo{person}{Andreas Geiger}.} \bibinfo{year}{2019}\natexlab{}.
\newblock \showarticletitle{Occupancy Networks: Learning 3D Reconstruction in Function Space}. In \bibinfo{booktitle}{\emph{CVPR}}.
\newblock


\bibitem[Mildenhall et~al\mbox{.}(2021)]%
        {mildenhall2021nerf}
\bibfield{author}{\bibinfo{person}{Ben Mildenhall}, \bibinfo{person}{Pratul~P Srinivasan}, \bibinfo{person}{Matthew Tancik}, \bibinfo{person}{Jonathan~T Barron}, \bibinfo{person}{Ravi Ramamoorthi}, {and} \bibinfo{person}{Ren Ng}.} \bibinfo{year}{2021}\natexlab{}.
\newblock \showarticletitle{Nerf: Representing scenes as neural radiance fields for view synthesis}.
\newblock \bibinfo{journal}{\emph{Commun. ACM}} \bibinfo{volume}{65}, \bibinfo{number}{1} (\bibinfo{year}{2021}), \bibinfo{pages}{99--106}.
\newblock


\bibitem[mkkellogg(2024)]%
        {mkkellogg2024GaussianSplats3D}
\bibfield{author}{\bibinfo{person}{mkkellogg}.} \bibinfo{year}{2024}\natexlab{}.
\newblock \bibinfo{title}{GaussianSplats3D}.
\newblock \bibinfo{howpublished}{\url{https://github.com/mkkellogg/GaussianSplats3D}}.
\newblock


\bibitem[M{\"u}ller(2006)]%
        {muller2006spherical}
\bibfield{author}{\bibinfo{person}{Claus M{\"u}ller}.} \bibinfo{year}{2006}\natexlab{}.
\newblock \bibinfo{booktitle}{\emph{Spherical harmonics}}. Vol.~\bibinfo{volume}{17}.
\newblock \bibinfo{publisher}{Springer}.
\newblock


\bibitem[M{\"u}ller et~al\mbox{.}(2022)]%
        {muller2022instant}
\bibfield{author}{\bibinfo{person}{Thomas M{\"u}ller}, \bibinfo{person}{Alex Evans}, \bibinfo{person}{Christoph Schied}, {and} \bibinfo{person}{Alexander Keller}.} \bibinfo{year}{2022}\natexlab{}.
\newblock \showarticletitle{Instant neural graphics primitives with a multiresolution hash encoding}.
\newblock \bibinfo{journal}{\emph{ACM transactions on graphics (TOG)}} (\bibinfo{year}{2022}).
\newblock


\bibitem[Newcombe et~al\mbox{.}(2015)]%
        {newcombe2015dynamicfusion}
\bibfield{author}{\bibinfo{person}{Richard~A Newcombe}, \bibinfo{person}{Dieter Fox}, {and} \bibinfo{person}{Steven~M Seitz}.} \bibinfo{year}{2015}\natexlab{}.
\newblock \showarticletitle{Dynamicfusion: Reconstruction and tracking of non-rigid scenes in real-time}.
\newblock


\bibitem[Park et~al\mbox{.}(2019)]%
        {park2019deepsdf}
\bibfield{author}{\bibinfo{person}{Jeong~Joon Park}, \bibinfo{person}{Peter Florence}, \bibinfo{person}{Julian Straub}, \bibinfo{person}{Richard Newcombe}, {and} \bibinfo{person}{Steven Lovegrove}.} \bibinfo{year}{2019}\natexlab{}.
\newblock \showarticletitle{Deepsdf: Learning continuous signed distance functions for shape representation}. In \bibinfo{booktitle}{\emph{Proceedings of the IEEE/CVF conference on computer vision and pattern recognition}}. \bibinfo{pages}{165--174}.
\newblock


\bibitem[Park et~al\mbox{.}(2021a)]%
        {park2021nerfies}
\bibfield{author}{\bibinfo{person}{Keunhong Park}, \bibinfo{person}{Utkarsh Sinha}, \bibinfo{person}{Jonathan~T Barron}, \bibinfo{person}{Sofien Bouaziz}, \bibinfo{person}{Dan~B Goldman}, \bibinfo{person}{Steven~M Seitz}, {and} \bibinfo{person}{Ricardo Martin-Brualla}.} \bibinfo{year}{2021}\natexlab{a}.
\newblock \showarticletitle{Nerfies: Deformable neural radiance fields}. In \bibinfo{booktitle}{\emph{Proceedings of the IEEE/CVF International Conference on Computer Vision}}. \bibinfo{pages}{5865--5874}.
\newblock


\bibitem[Park et~al\mbox{.}(2021b)]%
        {park2021hypernerf}
\bibfield{author}{\bibinfo{person}{Keunhong Park}, \bibinfo{person}{Utkarsh Sinha}, \bibinfo{person}{Peter Hedman}, \bibinfo{person}{Jonathan~T Barron}, \bibinfo{person}{Sofien Bouaziz}, \bibinfo{person}{Dan~B Goldman}, \bibinfo{person}{Ricardo Martin-Brualla}, {and} \bibinfo{person}{Steven~M Seitz}.} \bibinfo{year}{2021}\natexlab{b}.
\newblock \showarticletitle{Hypernerf: A higher-dimensional representation for topologically varying neural radiance fields}.
\newblock \bibinfo{journal}{\emph{arXiv preprint arXiv:2106.13228}} (\bibinfo{year}{2021}).
\newblock


\bibitem[Paszke et~al\mbox{.}(2019)]%
        {paszke2019pytorch}
\bibfield{author}{\bibinfo{person}{Adam Paszke}, \bibinfo{person}{Sam Gross}, \bibinfo{person}{Francisco Massa}, \bibinfo{person}{Adam Lerer}, \bibinfo{person}{James Bradbury}, \bibinfo{person}{Gregory Chanan}, \bibinfo{person}{Trevor Killeen}, \bibinfo{person}{Zeming Lin}, \bibinfo{person}{Natalia Gimelshein}, \bibinfo{person}{Luca Antiga}, \bibinfo{person}{Alban Desmaison}, \bibinfo{person}{Andreas Kopf}, \bibinfo{person}{Edward Yang}, \bibinfo{person}{Zachary DeVito}, \bibinfo{person}{Martin Raison}, \bibinfo{person}{Alykhan Tejani}, \bibinfo{person}{Sasank Chilamkurthy}, \bibinfo{person}{Benoit Steiner}, \bibinfo{person}{Lu Fang}, \bibinfo{person}{Junjie Bai}, {and} \bibinfo{person}{Soumith Chintala}.} \bibinfo{year}{2019}\natexlab{}.
\newblock \showarticletitle{PyTorch: An Imperative Style, High-Performance Deep Learning Library}. In \bibinfo{booktitle}{\emph{NeurIPS}}.
\newblock


\bibitem[Pumarola et~al\mbox{.}(2021a)]%
        {pumarola2021dnerf}
\bibfield{author}{\bibinfo{person}{Albert Pumarola}, \bibinfo{person}{Enric Corona}, \bibinfo{person}{Gerard Pons-Moll}, {and} \bibinfo{person}{Francesc Moreno-Noguer}.} \bibinfo{year}{2021}\natexlab{a}.
\newblock \showarticletitle{D-nerf: Neural radiance fields for dynamic scenes}. In \bibinfo{booktitle}{\emph{Proceedings of the IEEE/CVF Conference on Computer Vision and Pattern Recognition}}. \bibinfo{pages}{10318--10327}.
\newblock


\bibitem[Pumarola et~al\mbox{.}(2021b)]%
        {pumarola2021d}
\bibfield{author}{\bibinfo{person}{Albert Pumarola}, \bibinfo{person}{Enric Corona}, \bibinfo{person}{Gerard Pons-Moll}, {and} \bibinfo{person}{Francesc Moreno-Noguer}.} \bibinfo{year}{2021}\natexlab{b}.
\newblock \showarticletitle{D-nerf: Neural radiance fields for dynamic scenes}.
\newblock


\bibitem[Rakhimov et~al\mbox{.}(2022)]%
        {rakhimov2022npbg++}
\bibfield{author}{\bibinfo{person}{Ruslan Rakhimov}, \bibinfo{person}{Andrei-Timotei Ardelean}, \bibinfo{person}{Victor Lempitsky}, {and} \bibinfo{person}{Evgeny Burnaev}.} \bibinfo{year}{2022}\natexlab{}.
\newblock \showarticletitle{Npbg++: Accelerating neural point-based graphics}. In \bibinfo{booktitle}{\emph{CVPR}}.
\newblock


\bibitem[Ren et~al\mbox{.}(2023)]%
        {ren2023dreamgaussian4d}
\bibfield{author}{\bibinfo{person}{Jiawei Ren}, \bibinfo{person}{Liang Pan}, \bibinfo{person}{Jiaxiang Tang}, \bibinfo{person}{Chi Zhang}, \bibinfo{person}{Ang Cao}, \bibinfo{person}{Gang Zeng}, {and} \bibinfo{person}{Ziwei Liu}.} \bibinfo{year}{2023}\natexlab{}.
\newblock \showarticletitle{Dreamgaussian4d: Generative 4d gaussian splatting}.
\newblock \bibinfo{journal}{\emph{arXiv preprint arXiv:2312.17142}} (\bibinfo{year}{2023}).
\newblock


\bibitem[Rombach et~al\mbox{.}(2022)]%
        {ldm}
\bibfield{author}{\bibinfo{person}{Robin Rombach}, \bibinfo{person}{Andreas Blattmann}, \bibinfo{person}{Dominik Lorenz}, \bibinfo{person}{Patrick Esser}, {and} \bibinfo{person}{Bj\"orn Ommer}.} \bibinfo{year}{2022}\natexlab{}.
\newblock \showarticletitle{High-Resolution Image Synthesis With Latent Diffusion Models}. In \bibinfo{booktitle}{\emph{Proceedings of the IEEE/CVF Conference on Computer Vision and Pattern Recognition (CVPR)}}.
\newblock


\bibitem[R{\"u}ckert et~al\mbox{.}(2022)]%
        {ruckert2022adop}
\bibfield{author}{\bibinfo{person}{Darius R{\"u}ckert}, \bibinfo{person}{Linus Franke}, {and} \bibinfo{person}{Marc Stamminger}.} \bibinfo{year}{2022}\natexlab{}.
\newblock \showarticletitle{Adop: Approximate differentiable one-pixel point rendering}.
\newblock \bibinfo{journal}{\emph{ACM Transactions on Graphics (ToG)}} (\bibinfo{year}{2022}).
\newblock


\bibitem[Sanders and Kandrot(2010)]%
        {sanders2010cuda}
\bibfield{author}{\bibinfo{person}{Jason Sanders} {and} \bibinfo{person}{Edward Kandrot}.} \bibinfo{year}{2010}\natexlab{}.
\newblock \bibinfo{booktitle}{\emph{CUDA by example: an introduction to general-purpose GPU programming}}.
\newblock \bibinfo{publisher}{Addison-Wesley Professional}.
\newblock


\bibitem[Schonberger and Frahm(2016)]%
        {schonberger2016structure}
\bibfield{author}{\bibinfo{person}{Johannes~L Schonberger} {and} \bibinfo{person}{Jan-Michael Frahm}.} \bibinfo{year}{2016}\natexlab{}.
\newblock \showarticletitle{Structure-from-motion revisited}. In \bibinfo{booktitle}{\emph{Proceedings of the IEEE conference on computer vision and pattern recognition}}. \bibinfo{pages}{4104--4113}.
\newblock


\bibitem[Shreiner et~al\mbox{.}(2009)]%
        {shreiner2009opengl}
\bibfield{author}{\bibinfo{person}{Dave Shreiner} {et~al\mbox{.}}} \bibinfo{year}{2009}\natexlab{}.
\newblock \bibinfo{booktitle}{\emph{OpenGL programming guide: the official guide to learning OpenGL, versions 3.0 and 3.1}}.
\newblock \bibinfo{publisher}{Pearson Education}.
\newblock


\bibitem[Takikawa et~al\mbox{.}(2022)]%
        {takikawa2022variable}
\bibfield{author}{\bibinfo{person}{Towaki Takikawa}, \bibinfo{person}{Alex Evans}, \bibinfo{person}{Jonathan Tremblay}, \bibinfo{person}{Thomas M{\"u}ller}, \bibinfo{person}{Morgan McGuire}, \bibinfo{person}{Alec Jacobson}, {and} \bibinfo{person}{Sanja Fidler}.} \bibinfo{year}{2022}\natexlab{}.
\newblock \showarticletitle{Variable bitrate neural fields}. In \bibinfo{booktitle}{\emph{ACM SIGGRAPH 2022 Conference Proceedings}}.
\newblock


\bibitem[Tang et~al\mbox{.}(2022)]%
        {tang2022compressible}
\bibfield{author}{\bibinfo{person}{Jiaxiang Tang}, \bibinfo{person}{Xiaokang Chen}, \bibinfo{person}{Jingbo Wang}, {and} \bibinfo{person}{Gang Zeng}.} \bibinfo{year}{2022}\natexlab{}.
\newblock \showarticletitle{Compressible-composable nerf via rank-residual decomposition}.
\newblock \bibinfo{journal}{\emph{NeuRIPS}} (\bibinfo{year}{2022}).
\newblock


\bibitem[Tretschk et~al\mbox{.}(2021a)]%
        {tretschk2021non}
\bibfield{author}{\bibinfo{person}{Edgar Tretschk}, \bibinfo{person}{Ayush Tewari}, \bibinfo{person}{Vladislav Golyanik}, \bibinfo{person}{Michael Zollh{\"o}fer}, \bibinfo{person}{Christoph Lassner}, {and} \bibinfo{person}{Christian Theobalt}.} \bibinfo{year}{2021}\natexlab{a}.
\newblock \showarticletitle{Non-rigid neural radiance fields: Reconstruction and novel view synthesis of a dynamic scene from monocular video}.
\newblock


\bibitem[Tretschk et~al\mbox{.}(2021b)]%
        {tretschk2021nonrigid}
\bibfield{author}{\bibinfo{person}{Edgar Tretschk}, \bibinfo{person}{Ayush Tewari}, \bibinfo{person}{Vladislav Golyanik}, \bibinfo{person}{Michael Zollh\"{o}fer}, \bibinfo{person}{Christoph Lassner}, {and} \bibinfo{person}{Christian Theobalt}.} \bibinfo{year}{2021}\natexlab{b}.
\newblock \showarticletitle{Non-Rigid Neural Radiance Fields: Reconstruction and Novel View Synthesis of a Dynamic Scene From Monocular Video}. In \bibinfo{booktitle}{\emph{{IEEE} International Conference on Computer Vision ({ICCV})}}. {IEEE}.
\newblock


\bibitem[Wang et~al\mbox{.}(2022)]%
        {Wang_2022_CVPR}
\bibfield{author}{\bibinfo{person}{Liao Wang}, \bibinfo{person}{Jiakai Zhang}, \bibinfo{person}{Xinhang Liu}, \bibinfo{person}{Fuqiang Zhao}, \bibinfo{person}{Yanshun Zhang}, \bibinfo{person}{Yingliang Zhang}, \bibinfo{person}{Minye Wu}, \bibinfo{person}{Jingyi Yu}, {and} \bibinfo{person}{Lan Xu}.} \bibinfo{year}{2022}\natexlab{}.
\newblock \showarticletitle{Fourier PlenOctrees for Dynamic Radiance Field Rendering in Real-Time}. In \bibinfo{booktitle}{\emph{Proceedings of the IEEE/CVF Conference on Computer Vision and Pattern Recognition (CVPR)}}. \bibinfo{pages}{13524--13534}.
\newblock


\bibitem[Wang et~al\mbox{.}(2021)]%
        {wang2021ibrnet}
\bibfield{author}{\bibinfo{person}{Qianqian Wang}, \bibinfo{person}{Zhicheng Wang}, \bibinfo{person}{Kyle Genova}, \bibinfo{person}{Pratul~P Srinivasan}, \bibinfo{person}{Howard Zhou}, \bibinfo{person}{Jonathan~T Barron}, \bibinfo{person}{Ricardo Martin-Brualla}, \bibinfo{person}{Noah Snavely}, {and} \bibinfo{person}{Thomas Funkhouser}.} \bibinfo{year}{2021}\natexlab{}.
\newblock \showarticletitle{Ibrnet: Learning multi-view image-based rendering}.
\newblock


\bibitem[Wang et~al\mbox{.}(2004a)]%
        {wang2004image}
\bibfield{author}{\bibinfo{person}{Zhou Wang}, \bibinfo{person}{Alan~C Bovik}, \bibinfo{person}{Hamid~R Sheikh}, {and} \bibinfo{person}{Eero~P Simoncelli}.} \bibinfo{year}{2004}\natexlab{a}.
\newblock \showarticletitle{Image quality assessment: from error visibility to structural similarity}.
\newblock \bibinfo{journal}{\emph{IEEE transactions on image processing}} \bibinfo{volume}{13}, \bibinfo{number}{4} (\bibinfo{year}{2004}), \bibinfo{pages}{600--612}.
\newblock


\bibitem[Wang et~al\mbox{.}(2004b)]%
        {ssim}
\bibfield{author}{\bibinfo{person}{Zhou Wang}, \bibinfo{person}{Alan~C Bovik}, \bibinfo{person}{Hamid~R Sheikh}, {and} \bibinfo{person}{Eero~P Simoncelli}.} \bibinfo{year}{2004}\natexlab{b}.
\newblock \showarticletitle{Image quality assessment: from error visibility to structural similarity}.
\newblock \bibinfo{journal}{\emph{IEEE transactions on image processing}} (\bibinfo{year}{2004}).
\newblock


\bibitem[Wu et~al\mbox{.}(2023b)]%
        {wu20234d}
\bibfield{author}{\bibinfo{person}{Guanjun Wu}, \bibinfo{person}{Taoran Yi}, \bibinfo{person}{Jiemin Fang}, \bibinfo{person}{Lingxi Xie}, \bibinfo{person}{Xiaopeng Zhang}, \bibinfo{person}{Wei Wei}, \bibinfo{person}{Wenyu Liu}, \bibinfo{person}{Qi Tian}, {and} \bibinfo{person}{Wang Xinggang}.} \bibinfo{year}{2023}\natexlab{b}.
\newblock \showarticletitle{4D Gaussian Splatting for Real-Time Dynamic Scene Rendering}.
\newblock \bibinfo{journal}{\emph{arXiv preprint arXiv:2310.08528}} (\bibinfo{year}{2023}).
\newblock


\bibitem[Wu et~al\mbox{.}(2020)]%
        {wu2020nhr}
\bibfield{author}{\bibinfo{person}{Minye Wu}, \bibinfo{person}{Yuehao Wang}, \bibinfo{person}{Qiang Hu}, {and} \bibinfo{person}{Jingyi Yu}.} \bibinfo{year}{2020}\natexlab{}.
\newblock \showarticletitle{Multi-view neural human rendering}. In \bibinfo{booktitle}{\emph{Proceedings of the IEEE/CVF Conference on Computer Vision and Pattern Recognition}}. \bibinfo{pages}{1682--1691}.
\newblock


\bibitem[Wu et~al\mbox{.}(2023a)]%
        {wu2023reconfusion}
\bibfield{author}{\bibinfo{person}{Rundi Wu}, \bibinfo{person}{Ben Mildenhall}, \bibinfo{person}{Philipp Henzler}, \bibinfo{person}{Keunhong Park}, \bibinfo{person}{Ruiqi Gao}, \bibinfo{person}{Daniel Watson}, \bibinfo{person}{Pratul~P Srinivasan}, \bibinfo{person}{Dor Verbin}, \bibinfo{person}{Jonathan~T Barron}, \bibinfo{person}{Ben Poole}, {et~al\mbox{.}}} \bibinfo{year}{2023}\natexlab{a}.
\newblock \showarticletitle{Reconfusion: 3d reconstruction with diffusion priors}.
\newblock \bibinfo{journal}{\emph{arXiv preprint arXiv:2312.02981}} (\bibinfo{year}{2023}).
\newblock


\bibitem[Xu et~al\mbox{.}(2024a)]%
        {xu2024relightable}
\bibfield{author}{\bibinfo{person}{Zhen Xu}, \bibinfo{person}{Sida Peng}, \bibinfo{person}{Chen Geng}, \bibinfo{person}{Linzhan Mou}, \bibinfo{person}{Zihan Yan}, \bibinfo{person}{Jiaming Sun}, \bibinfo{person}{Hujun Bao}, {and} \bibinfo{person}{Xiaowei Zhou}.} \bibinfo{year}{2024}\natexlab{a}.
\newblock \showarticletitle{Relightable and Animatable Neural Avatar from Sparse-View Video}. In \bibinfo{booktitle}{\emph{CVPR}}.
\newblock


\bibitem[Xu et~al\mbox{.}(2024b)]%
        {xu20244k4d}
\bibfield{author}{\bibinfo{person}{Zhen Xu}, \bibinfo{person}{Sida Peng}, \bibinfo{person}{Haotong Lin}, \bibinfo{person}{Guangzhao He}, \bibinfo{person}{Jiaming Sun}, \bibinfo{person}{Yujun Shen}, \bibinfo{person}{Hujun Bao}, {and} \bibinfo{person}{Xiaowei Zhou}.} \bibinfo{year}{2024}\natexlab{b}.
\newblock \showarticletitle{4K4D: Real-Time 4D View Synthesis at 4K Resolution}. In \bibinfo{booktitle}{\emph{CVPR}}.
\newblock


\bibitem[Xu et~al\mbox{.}(2023)]%
        {xu2023easyvolcap}
\bibfield{author}{\bibinfo{person}{Zhen Xu}, \bibinfo{person}{Tao Xie}, \bibinfo{person}{Sida Peng}, \bibinfo{person}{Haotong Lin}, \bibinfo{person}{Qing Shuai}, \bibinfo{person}{Zhiyuan Yu}, \bibinfo{person}{Guangzhao He}, \bibinfo{person}{Jiaming Sun}, \bibinfo{person}{Hujun Bao}, {and} \bibinfo{person}{Xiaowei Zhou}.} \bibinfo{year}{2023}\natexlab{}.
\newblock \showarticletitle{EasyVolcap: Accelerating Neural Volumetric Video Research}.
\newblock  (\bibinfo{year}{2023}).
\newblock


\bibitem[Yang et~al\mbox{.}(2023a)]%
        {yang2023deformable}
\bibfield{author}{\bibinfo{person}{Ziyi Yang}, \bibinfo{person}{Xinyu Gao}, \bibinfo{person}{Wen Zhou}, \bibinfo{person}{Shaohui Jiao}, \bibinfo{person}{Yuqing Zhang}, {and} \bibinfo{person}{Xiaogang Jin}.} \bibinfo{year}{2023}\natexlab{a}.
\newblock \showarticletitle{Deformable 3D Gaussians for High-Fidelity Monocular Dynamic Scene Reconstruction}.
\newblock \bibinfo{journal}{\emph{arXiv preprint arXiv:2309.13101}} (\bibinfo{year}{2023}).
\newblock


\bibitem[Yang et~al\mbox{.}(2023b)]%
        {yang2023realtime}
\bibfield{author}{\bibinfo{person}{Zeyu Yang}, \bibinfo{person}{Hongye Yang}, \bibinfo{person}{Zijie Pan}, \bibinfo{person}{Xiatian Zhu}, {and} \bibinfo{person}{Li Zhang}.} \bibinfo{year}{2023}\natexlab{b}.
\newblock \showarticletitle{Real-time Photorealistic Dynamic Scene Representation and Rendering with 4D Gaussian Splatting}.
\newblock \bibinfo{journal}{\emph{arXiv preprint arXiv 2310.10642}} (\bibinfo{year}{2023}).
\newblock


\bibitem[Yu et~al\mbox{.}(2021)]%
        {yu2021pixelnerf}
\bibfield{author}{\bibinfo{person}{Alex Yu}, \bibinfo{person}{Vickie Ye}, \bibinfo{person}{Matthew Tancik}, {and} \bibinfo{person}{Angjoo Kanazawa}.} \bibinfo{year}{2021}\natexlab{}.
\newblock \showarticletitle{pixelnerf: Neural radiance fields from one or few images}.
\newblock


\bibitem[Yu et~al\mbox{.}(2017)]%
        {yu2017bodyfusion}
\bibfield{author}{\bibinfo{person}{Tao Yu}, \bibinfo{person}{Kaiwen Guo}, \bibinfo{person}{Feng Xu}, \bibinfo{person}{Yuan Dong}, \bibinfo{person}{Zhaoqi Su}, \bibinfo{person}{Jianhui Zhao}, \bibinfo{person}{Jianguo Li}, \bibinfo{person}{Qionghai Dai}, {and} \bibinfo{person}{Yebin Liu}.} \bibinfo{year}{2017}\natexlab{}.
\newblock \showarticletitle{Bodyfusion: Real-time capture of human motion and surface geometry using a single depth camera}.
\newblock


\bibitem[Yu et~al\mbox{.}(2018)]%
        {yu2018doublefusion}
\bibfield{author}{\bibinfo{person}{Tao Yu}, \bibinfo{person}{Zerong Zheng}, \bibinfo{person}{Kaiwen Guo}, \bibinfo{person}{Jianhui Zhao}, \bibinfo{person}{Qionghai Dai}, \bibinfo{person}{Hao Li}, \bibinfo{person}{Gerard Pons-Moll}, {and} \bibinfo{person}{Yebin Liu}.} \bibinfo{year}{2018}\natexlab{}.
\newblock \showarticletitle{Doublefusion: Real-time capture of human performances with inner body shapes from a single depth sensor}.
\newblock


\bibitem[Zhang et~al\mbox{.}(2018a)]%
        {zhang2018unreasonable}
\bibfield{author}{\bibinfo{person}{Richard Zhang}, \bibinfo{person}{Phillip Isola}, \bibinfo{person}{Alexei~A Efros}, \bibinfo{person}{Eli Shechtman}, {and} \bibinfo{person}{Oliver Wang}.} \bibinfo{year}{2018}\natexlab{a}.
\newblock \showarticletitle{The unreasonable effectiveness of deep features as a perceptual metric}. In \bibinfo{booktitle}{\emph{Proceedings of the IEEE conference on computer vision and pattern recognition}}. \bibinfo{pages}{586--595}.
\newblock


\bibitem[Zhang et~al\mbox{.}(2018b)]%
        {lpips}
\bibfield{author}{\bibinfo{person}{Richard Zhang}, \bibinfo{person}{Phillip Isola}, \bibinfo{person}{Alexei~A Efros}, \bibinfo{person}{Eli Shechtman}, {and} \bibinfo{person}{Oliver Wang}.} \bibinfo{year}{2018}\natexlab{b}.
\newblock \showarticletitle{The Unreasonable Effectiveness of Deep Features as a Perceptual Metric}. In \bibinfo{booktitle}{\emph{CVPR}}.
\newblock


\end{thebibliography}
\appendix

\end{document}